\pdfoutput=1

\documentclass[10pt,twocolumn,letterpaper]{article}

\usepackage{iccv}
\usepackage{times}
\usepackage{epsfig}
\usepackage{graphicx}
\usepackage{amsmath}
\usepackage{amssymb}
\usepackage{eucal}

\usepackage[bf,footnotesize]{caption}

\usepackage{cite}

\usepackage{cs}
\usepackage{mathsymb}
\usepackage{verbatim}
\usepackage{subfigure}
\usepackage{url}
\usepackage{graphicx}
\graphicspath{{./}{./Fig/}{./Figs/}{./Fig/eps/}{./Fig/pdf/}}

\usepackage{algorithm2e}

\let\savedalgorithm\algorithm
\let\savedendalgorithm\endalgorithm
\newenvironment{algorithmic}{%
\savedalgorithm
}{%
\savedendalgorithm
}

\usepackage{algorithm}
\usepackage[pagebackref=true,breaklinks=true,letterpaper=true,colorlinks,bookmarks=false]{hyperref}

\usepackage{grffile} %
\usepackage{slashbox,pict2e}

\usepackage{amsmath}
\usepackage{url}
\usepackage{multirow}

\def\Integer{\mathbb{Z}^+}

\iccvfinalcopy

\def\iccvPaperID{1624} %

\def\cone{{\ding{172}}}
\def\ctwo{{\ding{173}}}
\def\cthree{{\ding{174}}}
\def\cfour{{\ding{175}}}
\def\cfive{{\ding{176}}}

\def\algoname{{\rm pAUCEns}\xspace}
\def\AUC{{\rm{AUC} }}
\def\pAUC{{\rm{pAUC} }}
\def\FPR{{\rm{FPR} }}
\def\FPPI{{\rm{FPPI} }}

\def\Real{\mathbb{R}}

\def\sH{\mathcal{H}}

\def\bxip{{\bx_i^{+}}}
\def\bxjm{{\bx_j^{-}}}

\def\iprime{{l}}

\def\deltapauc{{ \Delta_{(\alpha,\beta)} }}
\def\tmax{{t_{\mathrm{max}}}}

\def\bpi{{\boldsymbol \pi}}
\def\bPi{{\boldsymbol \Pi}}
\def\bPimn{{\bPi_{m,n}}}
\def\bpiast{{{\boldsymbol \pi}^{\ast}}}
\def\bpihat{{\hat{\bpi}}}

\def\bHp{{\bH_{+}}}
\def\bHm{{\bH_{-}}}
\def\bSp{{\bS_{+}}}
\def\bSm{{\bS_{-}}}

\def\jalpha{{j_{\alpha}}}
\def\jalphaone{{j_{\alpha}+1}}
\def\jbeta{{j_{\beta}}}
\def\jbetaone{{j_{\beta}+1}}

\def\bOne{{\bf 1}}

\newcommand{\bigO}{\ensuremath{\mathcal{O}}}

\def\MyTiny{\fontsize{6.5pt}{6.5pt} \selectfont}

\begin{document}

\title{Efficient pedestrian detection by directly optimizing
    \\ the partial area under the ROC curve\thanks{Appearing in International Conference in Computer
        Vision (ICCV), 2013, Sydney, Australia.
    This work was in part supported by ARC Future Fellowship FT120100969.
    }
}

\author{Sakrapee Paisitkriangkrai, Chunhua Shen\thanks{Corresponding author
    (e-mail: chunhua.shen@icloud.com).},
    Anton van den Hengel\\
The University of Adelaide, SA 5005, Australia}

\maketitle

\begin{abstract}

Many typical applications of object detection operate within a
prescribed false-positive range.  In this situation the performance of
a detector should be assessed on the basis of the area under the ROC
curve over that range, rather than over the full curve, as the
performance outside the range is irrelevant.
This measure is labelled as the
    partial area
    under the ROC curve (pAUC).
Effective cascade-based classification, for example, depends on
training node classifiers that achieve the maximal detection rate
    at a moderate false positive rate, e.g., around 40\% to 50\%.
    We propose a novel ensemble learning method which
    achieves a maximal detection rate at a user-defined range of false
    positive rates by directly optimizing the partial AUC using
    structured learning.
    By optimizing for different ranges of false positive rates,
    the proposed method can be used to train either a single strong
    classifier or a node classifier
    forming part of a cascade classifier.
    Experimental results on both synthetic and
    real-world data sets demonstrate the effectiveness of our approach,
and we show that it is possible to
train state-of-the-art pedestrian detectors using the proposed
    structured ensemble learning method.

\end{abstract}

\section{Introduction}

Object detection is one of several fundamental topics
in computer vision.
The task of object detection is to identify predefined objects
in a given images using knowledge gained through analysis of a set of labelled positive and negative exemplars.
Viola and Jones' face detection algorithm \cite{Viola2004Robust}
forms the basis of many of the
state-of-the-art real-time algorithms for object detection tasks.

The most commonly adopted evaluation method by which to compare the detection performance
of different algorithms is the Receiver Operating Characteristic (ROC) curve.
The curve illustrates the varying performance of a binary classifier
system as its discrimination threshold is altered.
In the face and human detection literature researchers are often
interested in the low false positive area of the ROC curve since this
region
characterizes
the performance needed for most
real-world vision applications.
This is due to the fact that object detection is
a highly asymmetric classification problem as there are
only ever a small number of
target objects among the millions of background patches
in a single test image.
A false positive rate of $10^{-3}$ per scanning window
would result in thousands of false positives %
in a single image, which is impractical for most applications.
For
many tasks, and particularly  human detection,
 researchers also report the partial area
under the ROC curve (pAUC), typically over the range $0.01$ and $1.0$
false positives per image \cite{Dollar2012Pedestrian}.
As the name implies, pAUC is calculated as the
area under the ROC curve between two specified
false positive rates (FPRs).
It summarizes the
practical
performance of a detector and often is the
primary performance measure of interest.

Although pAUC is {\em the metric of interest} that has been used to
evaluate detection performance,
Most classifiers %
do not
directly optimize this evaluation criterion, and
as a result,
often under-perform.
In this paper, we present a principled approach for learning an ensemble classifier
which directly optimizes the \emph{partial} area under the ROC curve,
where the range over which the area is calculated may be selected according to the desired application.
Built upon the structured learning framework,
we thus propose here a novel form of ensemble classifier which directly optimizes the partial AUC score, which we call \algoname.
As with all other boosting algorithms, our approach learns a predictor by
building an ensemble of weak classification rules in a greedy fashion.
It also relies on a sample re-weighting mechanism to pass the information
between each iteration.
However, unlike traditional boosting, at each iteration,
the proposed approach places a greater emphasis on
samples
which have the incorrect ordering\footnote{The positive sample
has an incorrect ordering if it is ranked below the negative sample.
In other words, we want all positive samples to be ranked above all
negative samples.}
to achieve
the {\em optimal partial AUC score}.
The result is the ensemble learning method which yields the scoring function
consistent with the correct relative ordering of positive and negative samples
and optimizes the partial AUC score in
a false positive rate range $[\alpha, \beta]$
where $0 \leq \alpha < \beta \leq 1$.

\paragraph{Main contributions}
(1) We propose a new ensemble learning approach which explicitly
optimizes the partial area under the ROC curve (pAUC) between any two given false positive rates.
The method is of particular interest in the wide variety of
applications where performance is most important over a particular
range within the ROC curve.
The approach shares  similarities with conventional boosting methods,
but differs significantly in that the proposed method optimizes
a multivariate performance measure using structured learning.
Our design is simple and a conventional boosting-based visual detector
can be transformed into a \algoname-based visual detector with very
few modifications to the existing code. Our approach is efficient
since it exploits both the efficient weak classifier training
and the efficient cutting plane solver for optimizing the
partial AUC score in the structural SVM setting.
(2) We show that our approach is more intuitive and simpler to use
than alternative algorithms, such as Asymmetric
AdaBoost \cite{Viola2002Fast} and Cost-Sensitive
AdaBoost \cite{Masnadi2011Cost}, where one needs to cross-validate the
asymmetric parameter from a fixed set of discrete points.
Furthermore, it is unclear how one would set the asymmetric parameter
in order to achieve a maximal pAUC score for a specified false
positive range.
To our knowledge, our approach is the first principled ensemble method
that directly optimizes the partial AUC in an arbitrary false positive range
$[\alpha, \beta]$.
(3) Experimental results on several data sets, especially on
challenging human detection data sets, demonstrate the
effectiveness of the proposed approach.
Our pedestrian detector performs better than or on par with the
state-of-the-art, despite the fact that our detector only uses two
standard low-level image features.

\paragraph{Related work}
Various ensemble classifiers have been proposed in the literature.
Of these AdaBoost is one the most well known
as it
has achieved tremendous success in computer vision
and machine learning applications.
In object detection, the cost of missing a true target is often
higher than the cost of a false positive.
Classifiers that are optimal under the symmetric cost,
and thus treat false positives and negatives equally,
cannot exploit this information.
Several cost sensitive learning algorithms, where the classifier weights a positive
class more heavily than a negative class, have thus been proposed.

Viola and Jones introduced the asymmetry property
in Asymetric AdaBoost (AsymBoost) \cite{Viola2002Fast}.
However, the authors reported that this asymmetry is immediately
absorbed by the first weak classifier.
Heuristics are then used to avoid this problem.
Peng \etal proposed a fully-corrective asymmetric boosting method
which does not have this problem \cite{AsymBoost2011Wang}.
Note that one needs to carefully cross-validate the asymmetric parameter
in order to achieve the desired result.
Masnadi-Shirazi and Vasconcelos \cite{Masnadi2011Cost} proposed a
cost-sensitive boosting algorithm based on the
statistical interpretation of boosting.
Their approach is to optimize
the cost-sensitive loss by means of gradient descent.
Shen \etal proposed LACBoost and FisherBoost to address this asymmetry issue in cascade classifiers
\cite{FisherBoost2013IJCV}.
Most works along this line
address the pAUC evaluation criterion {\em indirectly}.
In addition, one needs to carefully cross-validate the asymmetric parameter
in order to maximize the detection rate in a particular false
positive range.

Several algorithms that directly optimize the pAUC score
have been proposed in bioinformatics
\cite{Hsu2012Linear,Komori2010Boosting}.
Komori and Eguchi  optimize the pAUC
using boosting-based algorithms \cite{Komori2010Boosting}.
This algorithm is heuristic in nature.
Narasimhan and Agarwal develop a structural SVM based method which directly optimizes the pAUC score \cite{Narasimhan2013Structural}.
They demonstrate that their approach, which uses a support vector
method, significantly outperforms several existing algorithms,
including pAUCBoost \cite{Komori2010Boosting} and asymmetric SVM
\cite{Wu2008Asymmetric}.  Building on Narasimhan and Agarwal's work,
we propose the principled fully-corrective ensemble method which
directly optimizes the pAUC evaluation criterion.
The approach is flexible and can be applied to an arbitrary false
positive range $[\alpha, \beta]$.
To our knowledge, our approach is the first principled ensemble
learning method
that directly optimizes the partial AUC in a false positive range
not bounded by zero.
It is important to emphasize here the difference between our approach
and that of \cite{Narasimhan2013Structural}.
\cite{Narasimhan2013Structural} train a linear structural SVM
while our approach learns the ensemble of classifiers.
For pedestrian detection, HOG with the ensemble of classifiers reduces
the average miss-rate over HOG+SVM by more than $30\%$ \cite{Benenson2013Seeking}.

\paragraph{Notation}
Bold lower-case letters, \eg, $\bw$, denote column vectors and bold upper-case letters,
\eg, $\bH$, denote matrices.
Let $\{\bxip\}_{i=1}^{m}$ be the set of positive training data
and $\{\bxjm\}_{j=1}^{n}$ be the set of negative training data.
A set of all training samples can be written as
$\bS = (\bSp, \bSm)$ where
$\bSp = (\bx_{1}^{+}, \cdots, \bx_{m}^{+})$ and
$\bSm = (\bx_{1}^{-}, \cdots, \bx_{n}^{-})$.
We denote by $\sH$ a set of all possible outputs of weak learners.
Assuming that we have $k$ possible weak learners,
the output of weak learners for positive and negative data
can be represented as $\bH = (\bHp, \bHm)$
where $\bH_{+} \in \Real^{k \times m}$
and  $\bH_{-} \in \Real^{k \times n}$, respectively.
Here $h^{+}_{ti}$ is the label predicted by the weak learner
$\hbar_t(\cdot)$ on the positive training data $\bxip$.
Each column $\bh_{:l}$ of the matrix $\bH$ represents the output of all weak learners
when applied to the training instance $\bx_{l}$.
Each row $\bh_{t:}$ of the matrix $\bH$ represents the output predicted
by the weak learner $\hbar_t(\cdot)$ on all the training data.
The goal is to learn a set of binary weak learners and a scoring function,
$f: \Real^{k} \rightarrow \Real$,
that has good performance in terms of the pAUC between
some specified false positive rates $\alpha$ and $\beta$
where $0 \leq \alpha < \beta \leq 1$.

\paragraph{Structured learning approach for optimizing pAUC}
Before we propose our approach, we briefly review the concept of
SVM$_{\rm pAUC}$ $[\alpha,\beta]$ \cite{Narasimhan2013Structural},
in which our ensemble learning approach is built upon.
Unless otherwise stated, we follow the symbols used in \cite{Narasimhan2013Structural}.
The area under the empirical ROC curve (AUC) can be defined as,
\begin{align}
    \label{EQ:auc}
        \AUC = \frac{1}{mn} \sum_{i=1}^m \sum_{j=1}^n \bOne \bigl( f(\bxip) > f(\bxjm) \bigr),
\end{align}
and the partial AUC in the false positive range $\left[ \alpha, \beta \right]$
can be written as \cite{Dodd2003Partial, Narasimhan2013Structural},
\begin{align}
    \label{EQ:pauc}
        \notag
        \pAUC &= \frac{1}{mn(\beta-\alpha)} \sum_{i=1}^m
          \bigl( p_1 (\alpha) + p_2 (\alpha, \beta) + p_3(\beta) \bigr), \\ \notag
        p_1 (\alpha) &= (\jalpha - n \alpha) \cdot
            \bOne \bigl( f(\bxip) > f(\bx_{(\jalpha)}^-) \bigr), \\ \notag
        p_2 (\alpha,\beta) &= {\textstyle \sum}_{j=\jalphaone}^{\jbeta}
            \bOne \bigl( f(\bxip) > f(\bx_{(j)}^-) \bigr), \\
        p_3 (\beta) &= (n \beta - \jbeta) \cdot
            \bOne \bigl( f(\bxip) > f(\bx_{(\jbetaone)}^-) \bigr),
\end{align}
where $\jalpha = \lceil n \alpha \rceil$, $\jbeta = \lfloor n \beta \rfloor$,
$\bx_{(j)}^-$ denotes the negative instance in $\bSm$ ranked in the $j$-th position
amongst negative samples in descending order of scores.
$p_1(\alpha)$, $p_2(\alpha,\beta)$ and $p_3(\beta)$
correspond to the sum of detection rates at
$\FPR = \left[\alpha, \frac{\jalpha}{n} \right]$,
$\FPR = \left[ \frac{\jalpha}{n}, \frac{\jbeta}{n} \right]$,
and
$\FPR = \left[\frac{\jbeta}{n}, \beta \right]$, respectively.

Given a training sample $\bS = (\bSp, \bSm)$, our objective is to
find a linear function $\bw^\T \bx$
that optimizes the pAUC in an \FPR range of
$[\alpha, \beta]$.
We cast this pAUC optimization problem as a structural learning task.
For any ordering of the training instances, the relative ordering of $m$ positive instances and
$n$ negative instances is represented via a matrix $\bpi \in \{0,1\}^{m \times n}$ where,
\begin{align}
    \label{EQ:piij}
        \pi_{ij}  =
            \begin{cases}
                0  \quad& \text{if} \; \bxip \; \text{is ranked above} \; \bxjm  \\
                1  \quad& \text{otherwise.}
            \end{cases}
\end{align}
We define the correct relative ordering of $\bpi$ as $\bpiast$ where
$\pi^{\ast}_{ij} = 0, \forall i, j$.
The pAUC loss
in the false positive range $[\alpha, \beta]$ of $\bpi$
with respect to $\bpiast$ can be written as,
\begin{align}
    \label{EQ:deltapauc}
        \notag
        \deltapauc (\bpiast, &\bpi) = \frac{1}{mn(\beta-\alpha)} \sum_{i=1}^m
          \bigl[ (\jalpha - n \alpha) \pi_{i,(\jalpha)_{\bpi}} + \\
                & {\textstyle \sum}_{j=\jalphaone}^{\jbeta} \pi_{i,(j)_{\bpi}} +
                 (n \beta - \jbeta) \pi_{i,(\jbetaone)_{\bpi}} \bigr],
\end{align}
where $(j)_{\bpi}$ denotes the index of the negative instance
consistent with the matrix $\bpi$.
We define the joint feature map $\phi$ of the form
\begin{align}
    \label{EQ:featmap}
        \phi (\bS, \bpi) = \frac{1}{mn(\beta-\alpha)}
            {\textstyle \sum}_{i,j} (1 - \pi_{ij}) (\bxip - \bxjm).
\end{align}
The choice of $\phi(\bS, \bpi)$ over $\bpi \in \bPimn$
guarantees that the variable $\bw$, which optimizes
$\bw^\T \phi(\bS, \bpi)$, will also produce
the scoring function $f(\bx) = \bw^\T \bx$ that
achieves the optimal partial AUC score.
The above problem can be summarized as the following convex
optimization problem \cite{Narasimhan2013Structural}:
\begin{align}
\label{EQ:hinge1}
    \min_{ \bw , \xi }   \quad
    &
    \frac{1}{2} \| \bw  \|_{2}^{2} + \nu \, \xi    \\ \notag
    \st \; &
    \bw^\T ( \phi(\bS, \bpiast) - \phi(\bS, \bpi) )
    \geq \deltapauc (\bpi^{\ast}, \bpi) - \xi,
\end{align}
$\forall \bpi \in \bPimn$ and $\xi \geq 0$.
Note that $\bpiast$ denote the correct relative ordering
and $\bpi$ denote any arbitrary orderings.

\section{Our approach}

In order to design an ensemble-like algorithm for the pAUC,
we first introduce a projection function,
$\hbar(\cdot)$, which projects an instance vector
$\bx$ to $\{-1,+1\}$.
This projection function is also known as
the weak learner in boosting.
In contrast to the previously described structured learning,
we learn the scoring function, which optimizes the area
under the curve between two false positive rates
of the form:
$f(\bx) = \sum_{t=1}^k w_{t} \hbar_t(\bx)$
where $\bw \in \Real^{k}$ is the linear coefficient vector and
$\{\hbar_t(\cdot)\}_{t=1}^{k}$ denote a set of binary weak learners.
Let us assume that we have already learned a set of all projection functions.
By using the same pAUC loss, $\deltapauc(\cdot, \cdot)$, as in \eqref{EQ:deltapauc},
and the same feature mapping, $\phi(\cdot, \cdot)$, as in \eqref{EQ:featmap},
the optimization problem we want to solve is:
\begin{align}
\label{EQ:hinge1a}
    \min_{ \bw , \xi }   \quad
    &
    \frac{1}{2} \| \bw  \|_{2}^{2} + \nu \, \xi    \\ \notag
    \st \; &
    \bw^\T ( \phi(\bH, \bpiast) - \phi(\bH, \bpi) )
    \geq \deltapauc (\bpi^{\ast}, \bpi) - \xi,
\end{align}
$\forall \bpi \in \bPimn$ and $\xi \geq 0$.
$\bH = (\bHp, \bHm)$ is the projected output
for positive and negative training samples.
$\phi(\bH, \bpi) = [\phi(\bh_{1:}, \bpi), \cdots, \phi(\bh_{k:}, \bpi)]$
where $\phi (\bh_{t:}, \bpi): (\Real^{m} \times \Real^{n}) \times \bPimn \rightarrow \Real$
and it is defined as,
\begin{align}
    \label{EQ:featmap2}
    \phi (\bh_{t:}, \bpi) = \frac{1}{mn(\beta-\alpha)}
           {\textstyle \sum}_{i,j} (1 &- \pi_{ij}) \\ \notag
            &\bigl( \hbar_{t} (\bxip) - \hbar_{t} (\bxjm) \bigr).
\end{align}
The only difference between \eqref{EQ:hinge1} and \eqref{EQ:hinge1a}
is that the original data is now projected to a new non-linear feature space.
We will show how this can further improved the pAUC score
in the experiment section.
The dual problem of \eqref{EQ:hinge1a} can be written as (see supplementary),
\begin{align}
    \label{EQ:hinge2}
        \max_{ \blambda }   \quad &
        {\textstyle \sum_{\bpi}} \blambda_{(\bpi)} \deltapauc(\bpiast, \bpi) - \\ \notag
        &\quad \frac{1}{2}  {\textstyle \sum_{\bpi,\bpihat}}
            \blambda_{(\bpi)} \blambda_{(\bpihat)}
            \langle   \phi_{\Delta}(\bH, \bpi), \phi_{\Delta}(\bH, \bpihat) \rangle  \\ \notag
        \st \quad &
        0 \leq {\textstyle \sum_{\bpi}} \blambda (\bpi) \leq \nu.
\end{align}
where $\blambda$ is the dual variable,
$\blambda_{(\bpi)}$ denotes the dual variable
associated with the inequality constraint for $\bpi \in \bPimn$
and $\phi_{\Delta}(\bH, \bpi) = \phi(\bH, \bpiast)  - \phi(\bH, \bpi)$.
To derive the Lagrange dual problem, the following KKT condition
is used,
\begin{align}
    \label{EQ:KKT}
        \bw = \sum_{\bpi \in \bPimn} \blambda_{(\bpi)}
        \bigl( \phi(\bH, \bpiast)  - \phi(\bH, \bpi) \bigr).
\end{align}

\paragraph{Finding best weak learners}
In this section, we show how one can explicitly learn the projection
function, $\hbar(\cdot)$.
We use the idea of column generation to derive an ensemble-like
algorithm similar to LPBoost \cite{demiriz2002}.
The condition for applying the column generation is that the duality gap
between the primal and dual problem is zero (strong duality).
By inspecting the KKT condition,
at optimality, \eqref{EQ:KKT} must hold for all $t = 1, \cdots, k$.
In other words, $w_t = \sum_{\bpi \in \bPimn} \blambda_{(\bpi)}
\bigl( \phi(\bh_{t:}, \bpiast)  - \phi(\bh_{t:}, \bpi) \bigr)$
must hold for all $t$.

For the weak learner in the current working set, the corresponding
condition in \eqref{EQ:KKT} is satisfied by the current solution.
For the weak learner that are not yet selected,
they do not appear in the current restricted optimization problem
and the corresponding $w_t = 0$.
It is easy to see that if $\sum_{\bpi \in \bPimn} \blambda_{(\bpi)}
\bigl( \phi(\bh_{t:}, \bpiast)  - \phi(\bh_{t:}, \bpi) \bigr) = 0$
for any $\hbar_t(\cdot)$ that are not in the
current working set, then the current solution
is already the globally optimal one.
Hence the subproblem for selecting the best
weak learner is:
\begin{align}
    \label{EQ:weak1}
    \hbar^{\ast}(\cdot) = \argmax_{\hbar \in \sH }
    \;
    \Bigl| {\textstyle \sum}_{\bpi} \blambda_{(\bpi)}
        \bigl(  \phi(\bh, \bpiast) - \phi (\bh, \bpi) \bigr) \Bigr|.
\end{align}
In other words, we pick the weak learner with the
value $| {\textstyle \sum}_{\bpi} \blambda_{(\bpi)}
        \bigl(  \phi(\bh, \bpiast) - \phi (\bh, \bpi) \bigr) |$
most deviated from zero.
At iteration $t$, we pick the most optimal weak learner from $\sH$.
Substituting \eqref{EQ:featmap2} into \eqref{EQ:weak1},
the subproblem for generating the optimal weak learner at iteration $t$
can be defined as,
\begin{align}
    \label{EQ:weak2}
    \notag
    \hbar_t^{\ast}(\cdot) &= \argmax_{\hbar \in \sH}
    \;
    \Bigl| \sum_{\bpi} \blambda_{(\bpi)}
        \sum_{i,j} \pi_{ij} \bigl( \hbar(\bxip) - \hbar(\bxjm) \bigr) \Bigr| \\ \notag
    &= \argmax_{\hbar \in \sH}
    \;
    \Bigl| \sum_{i,j} \bigl( {\textstyle \sum}_{\bpi} \blambda_{(\bpi)} \pi_{ij} \bigr)
          \bigl( \hbar(\bxip) - \hbar(\bxjm) \bigr) \Bigr| \\ \notag
    &= \argmax_{\hbar \in \sH}
    \;
    \Bigl| {\textstyle \sum}_{ \iprime } u_{\iprime} y_{\iprime} \hbar (\bx_{\iprime}) \Bigr| \\
    &= \argmax_{\hbar \in \sH}
    \;
    {\textstyle \sum}_{ \iprime } u_{\iprime} y_{\iprime} \hbar (\bx_{\iprime})
\end{align}
where $i$, $j$, $\iprime$ index the positive training samples ($i = 1, \cdots, m$),
the negative training samples ($j = 1, \cdots, n$) and
the entire training samples ($\iprime = 1, 2$,$\cdots$,$m+n$), respectively.
Here
\begin{align}
    \label{EQ:updatew}
        u_{\iprime}  =
            \begin{cases}
                \sum_{\bpi,j} \blambda_{(\bpi)} \pi_{\iprime j}
                    \quad& \text{if} \; y_{\iprime} = +1  \\
                \sum_{\bpi,i} \blambda_{(\bpi)} \pi_{i \iprime}
                    \quad& \text{if} \; y_{\iprime} = -1.
            \end{cases}
\end{align}
For decision stumps, the last equation in \eqref{EQ:weak2} is always valid since
the weak learner set $\sH$ is negation-closed \cite{Komori2011Boosting}.
In other words, if $\hbar(\cdot) \in \sH$, then
$[-\hbar](\cdot) \in \sH$, and vice versa.
Here $[-\hbar](\cdot) = -\hbar(\cdot)$.
For decision stumps, one can flip the inequality sign such that
$\hbar(\cdot) \in \sH$ and $[-\hbar](\cdot) \in \sH$.
In fact, any linear classifiers of the form
$\sign(\sum_{t} a_t x_t + a_{0})$ are negation-closed.
Using \eqref{EQ:weak2} to choose the best weak learner is
not heuristic as the solution to \eqref{EQ:weak1} decreases
the duality gap the most for the current solution.
See supplementary for more details.

\paragraph{Optimizing weak learners' coefficients}
We solve for the optimal $\bw$ that minimizes our
objective function \eqref{EQ:hinge1a}.
However, the optimization problem \eqref{EQ:hinge1a} has
an exponential number of constraints, one for each matrix
$\bpi \in \bPimn$.
As in \cite{Narasimhan2013Structural, Joachims2009Cutting},
we use the cutting plane method to solve this problem.
The basic idea of the cutting plane is that a small subset
of the constraints are sufficient to find an $\epsilon$-approximate
solution to the original problem.
The algorithm starts with an empty constraint set and it adds the most
violated constraint set at each iteration.
The QP problem is solved using linear SVM and the process
continues until no constraint is violated by more than $\epsilon$.
Since, the quadratic program is of constant size and
the cutting plane method converges in a constant number of iterations,
the major bottleneck lies in the combinatorial optimization (over
$\bPimn$) associated with finding the most violated constraint set
at each iteration.
Narasimhan and Agarwal show how this combinatorial problem
can be solved efficiently in a polynomial time \cite{Narasimhan2013Structural}.
We briefly discuss their efficient algorithm in this section.

The combinatorial optimization problem associated
with finding the most violated constraint can be written as,
\begin{align}
    \label{EQ:optw1}
        \bar{\bpi} =
             \argmax_{\bpi \in \bPimn }
    \;
    &Q_{\bw} (\bpi),
\end{align}
where
\begin{align}
    \label{EQ:optw2}
        Q_{\bw} (\bpi) =
    &\deltapauc(\bpiast, \bpi) - \\ \notag
     &\frac{1}{mn(\beta-\alpha)}
         \sum_{i,j}  \pi_{ij} \bw^\T (\bh_{:i}^{+} - \bh_{:j}^{-} ).
\end{align}
The trick to speed up \eqref{EQ:optw1} is to
note that any ordering of the instances that is
consistent with $\bpi$ yields the same objective value,
$Q_{\bw} (\bpi)$ in \eqref{EQ:optw2}.
In addition, one can break down \eqref{EQ:optw1}
into smaller maximization problems
by restricting the search
space from $\bPimn$ to the set $\bPi_{m,n}^{\bw}$ where
\begin{align}
    \notag
        \bPi_{m,n}^{\bw} =  \bigl\{ \bpi \in \bPimn
           | \; \forall i, j_1 < j_2 : \pi_{i,(j_1)_\bw} \geq \pi_{i,(j_2)_\bw} \bigr\}.
\end{align}
Here $\bPi_{m,n}^{\bw}$ represents the set of all matrices $\bpi$
in which the ordering of the scores of two negative instances,
$\bw^\T \bh_{:j_1}^{-}$ and $\bw^\T \bh_{:j_2}^{-}$, is consistent.
The new optimization problem is now easier to solve
as the set of negative instances over which the loss term in
\eqref{EQ:optw2} is computed is the same for all
orderings in the search space.
This simplification allows one to reduce the computational complexity
of \eqref{EQ:optw2}
to $\bigO \bigl( (m+n) \log (m+n) \bigr)$.
Interested reader may refer to \cite{Narasimhan2013Structural}.

\SetKwInput{KwInit}{Initialize}

\begin{algorithm}[t]
\caption{The training algorithm for \algoname.
}
\begin{algorithmic}
\footnotesize{
   \KwIn{
    \\1)    A set of training examples $\{\bx_l,y_l\}$, $l=1, \cdots, m+n$;
     $
     \;
     $
     \\ 2)    The maximum number of weak learners, $\tmax$;
     $
     \;
     $
     \\ 3)    The regularization parameter, $\nu$;
     $
     \;
     $
     \\ 4)    The learning objective based on the partial AUC, $\alpha$ and $\beta$;
   }

\KwOut{
  The scoring function$^{\dag}$, $f(\bx) = \sum_{t=1}^{\tmax} w_t \hbar_t(\bx)$,
  that optimizes the \pAUC score in the \FPR range $[\alpha, \beta]$;
}

\KwInit {
\\1) $t = 0$;
\\2) Initilaize sample weights: $u_l = \frac{0.5}{m}$ if $y_l = +1$, else $u_l = \frac{0.5}{n}$;
\\3) Extract low level features and store them in the cache memory
for fast data access;
}

\While{ $t < \tmax$}
{
  \cone\ Train a new weak learner using \eqref{EQ:weak2}. The weak learner corresponds to
     the weak classifier with the minimal weighted error (maximal edge) ;
  \\ \ctwo\ Add the best weak learner into the current set;
  \\ \cthree\ Solve the structured SVM problem using the cutting plane
          algorithm \cite{Narasimhan2013Structural};
  \\ \cfour\ Update sample weights, $\bu$, using \eqref{EQ:updatew} ;
  \\ \cfive\ $t \leftarrow t + 1;$
}

$\dag$ For a node in a cascade classifier, we introduce the threshold,
$b$, and adjust $b$ using the validation set
such that $\sign \bigl( f(\bx) - b \bigr)$ achieves the node learning objective \;

} 
\end{algorithmic}
\label{ALG:main}
\end{algorithm}

\paragraph{Discussion}
Our final ensemble classifier has a similar form as the
AdaBoost-based object detector of \cite{Viola2004Robust}.
Based on Algorithm~\ref{ALG:main},
step \cone\ and \ctwo\ of our algorithm are exactly the same as \cite{Viola2004Robust}.
Similar to AdaBoost, $u_\iprime$ in step \cone\ plays the role of sample weights
associated to each training sample.
The major difference between AdaBoost and our approach is in step \cthree\ and \cfour\,
where the weak learner's coefficient is computed and the sample weights
are updated.
In AdaBoost, the weak learner's coefficient is calculated as
$w_t = \frac{1}{2} \log \frac{1-\epsilon_t}{\epsilon_t}$
where $\epsilon_t = \sum_{l} u_l I \bigl( y_l \neq \hbar_t(\bx_{l}) \bigr)$
and $I$ is the indicator function.
The sample weights are updated with
$u_l = \frac{ u_l \exp (-w_t y_l \hbar_t(\bx_{l}))}
{\sum_{l} u_l \exp (-w_t y_l \hbar_t(\bx_{l})}$.
We point this out here since a minimal modification is required in order
to transform the existing implementation of AdaBoost to \algoname.
Given the existing code of AdaBoost and
the publicly available implementation
of \cite{Narasimhan2013Structural},
our \algoname can be implemented in less than $10$ lines of codes.
A computational complexity analysis of our approach
can be found in the supplementary.

In the next section, we train two different types of classifiers: the
strong classifier \cite{Dollar2009Integral}
and the node classifier \cite{Viola2004Robust, Wu2008Fast}.
For the strong classifier, we set the value of $\alpha$ and $\beta$ based on
the evaluation criterion.
For the node classifier, we set the value of $\alpha$ and $\beta$ in each node to be
$0.49$ and $0.51$, respectively.

\section{Experiments}
 \begin{figure*}[t]
        \centering
        \includegraphics[width=0.23\textwidth,clip]{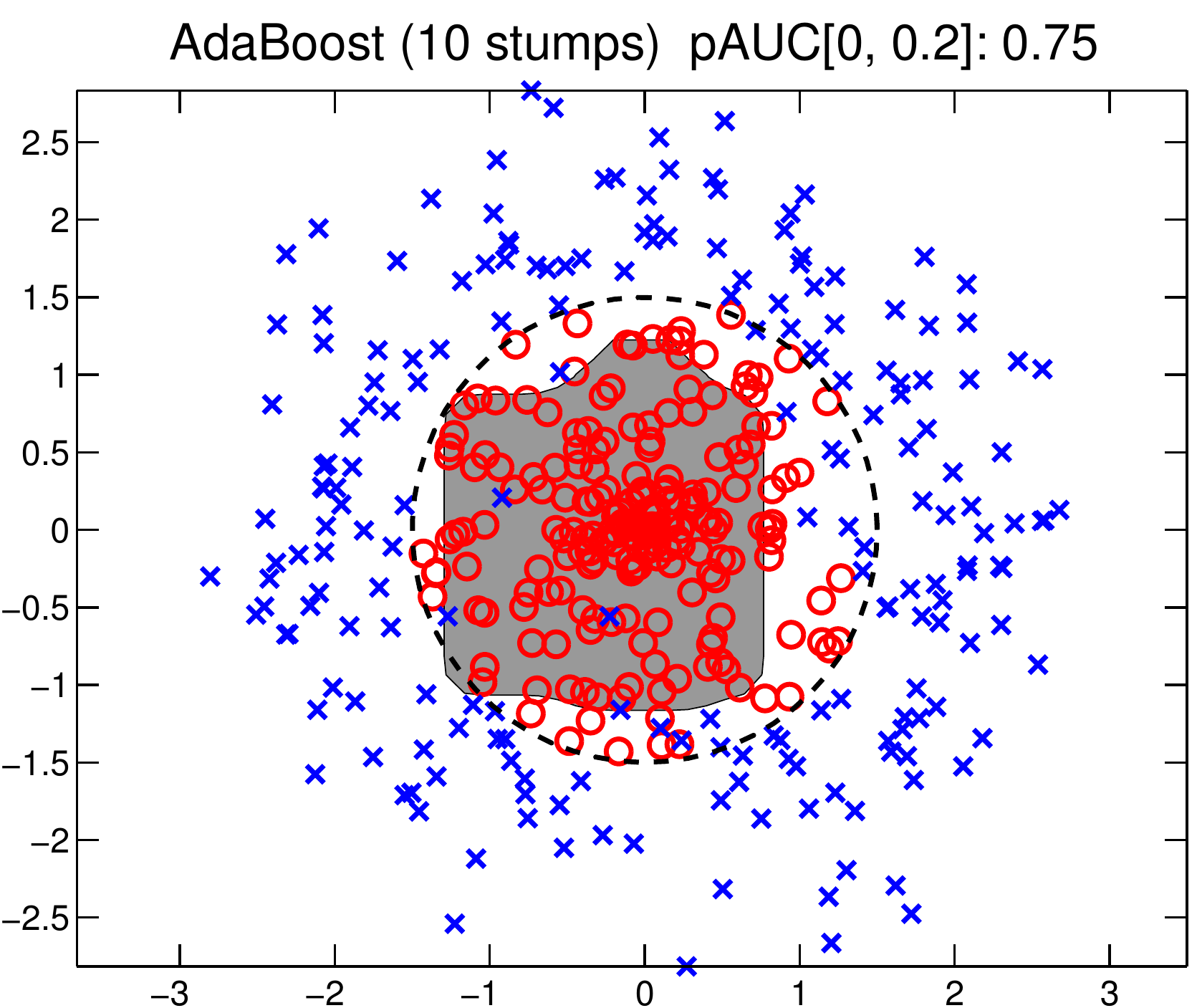}
        \includegraphics[width=0.23\textwidth,clip]{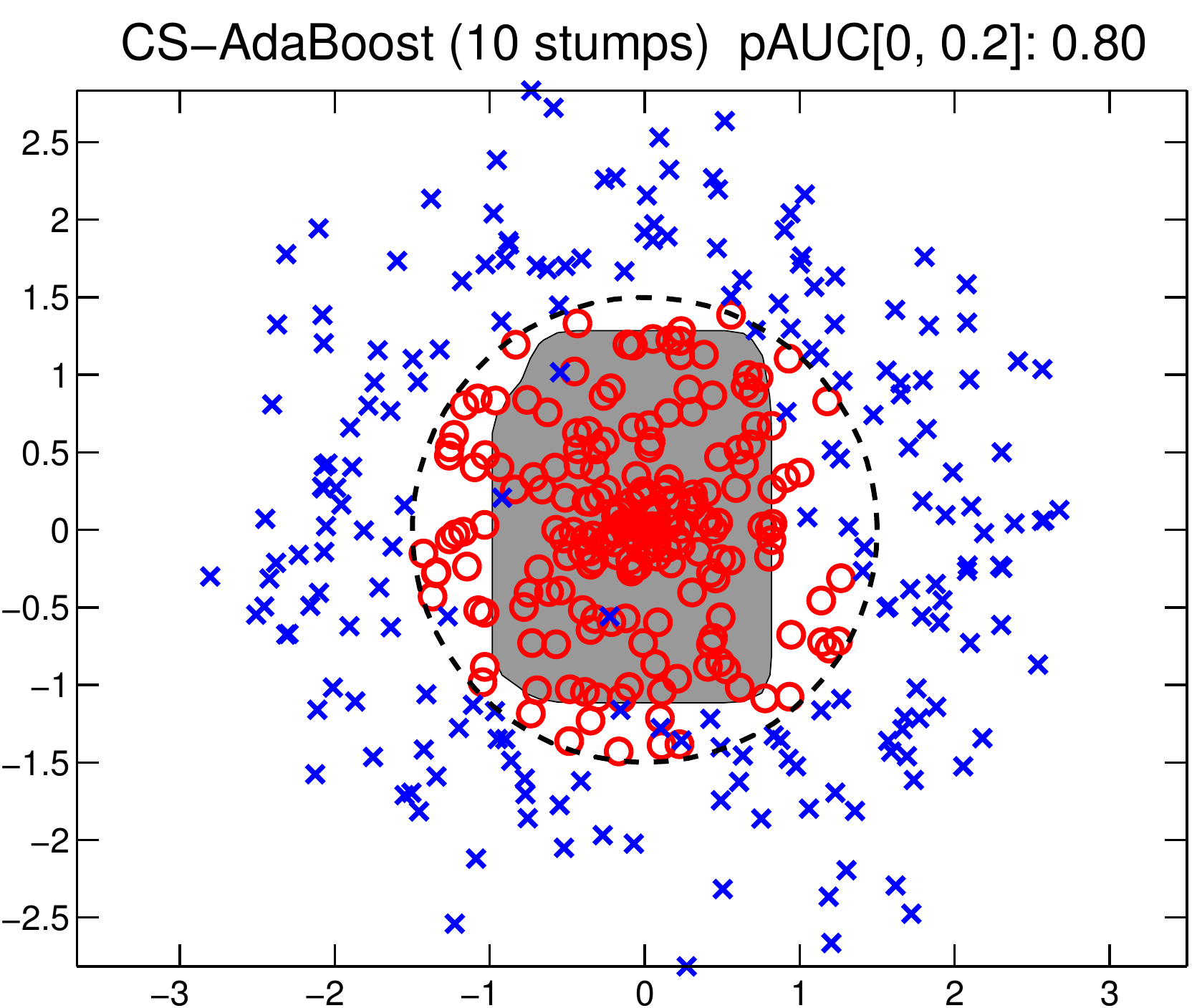}
        \includegraphics[width=0.23\textwidth,clip]{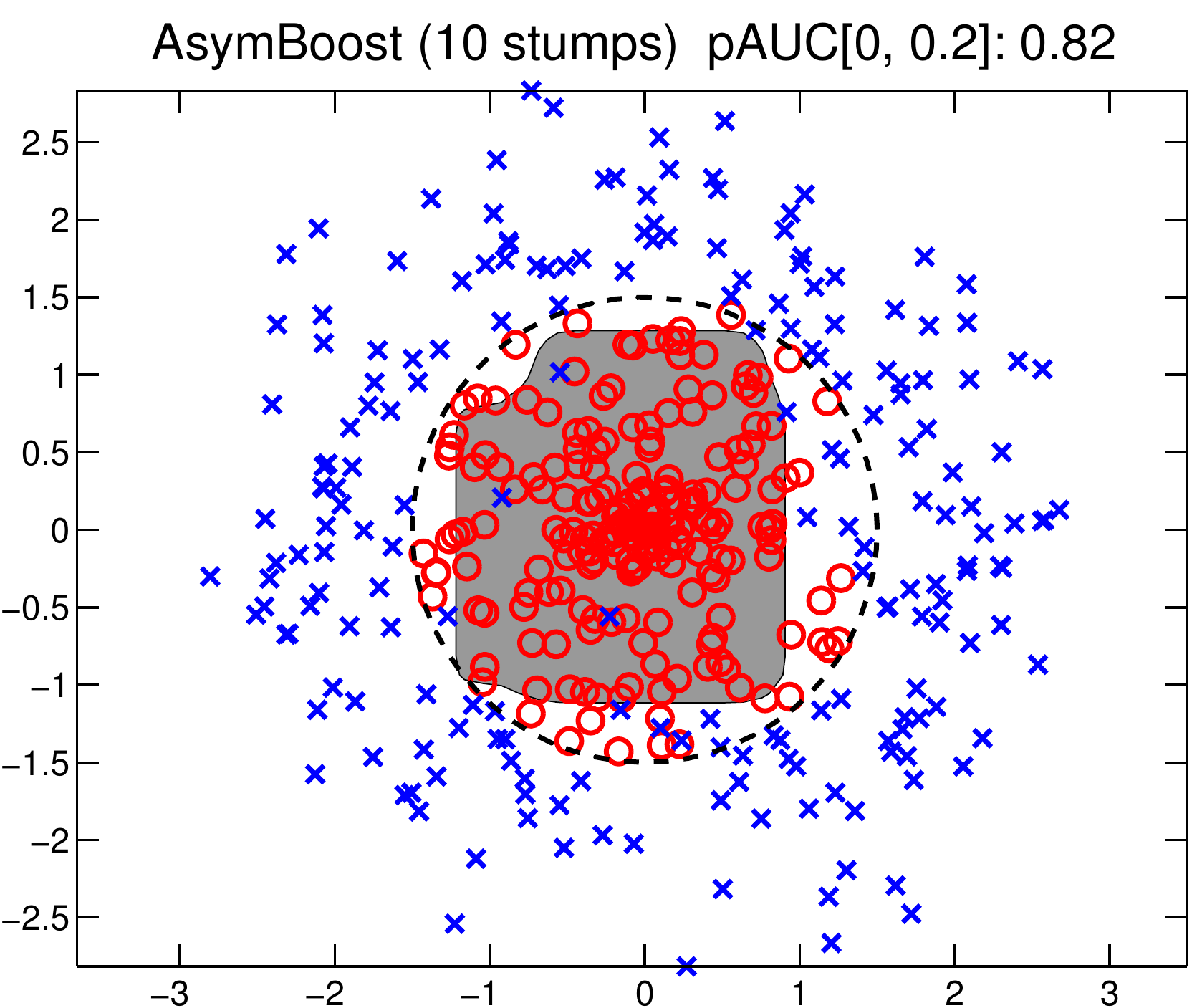}
        \includegraphics[width=0.23\textwidth,clip]{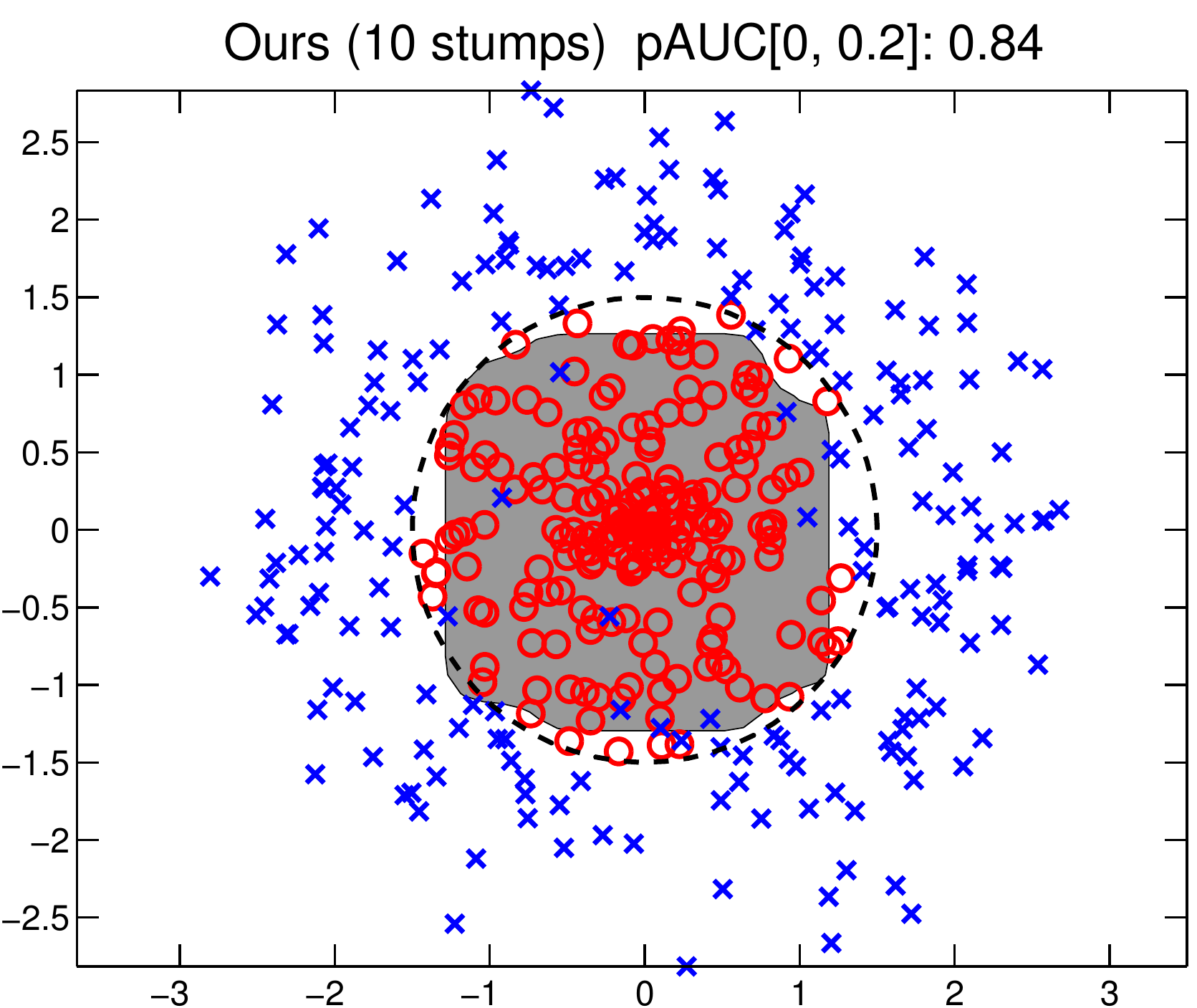}
        \includegraphics[width=0.23\textwidth,clip]{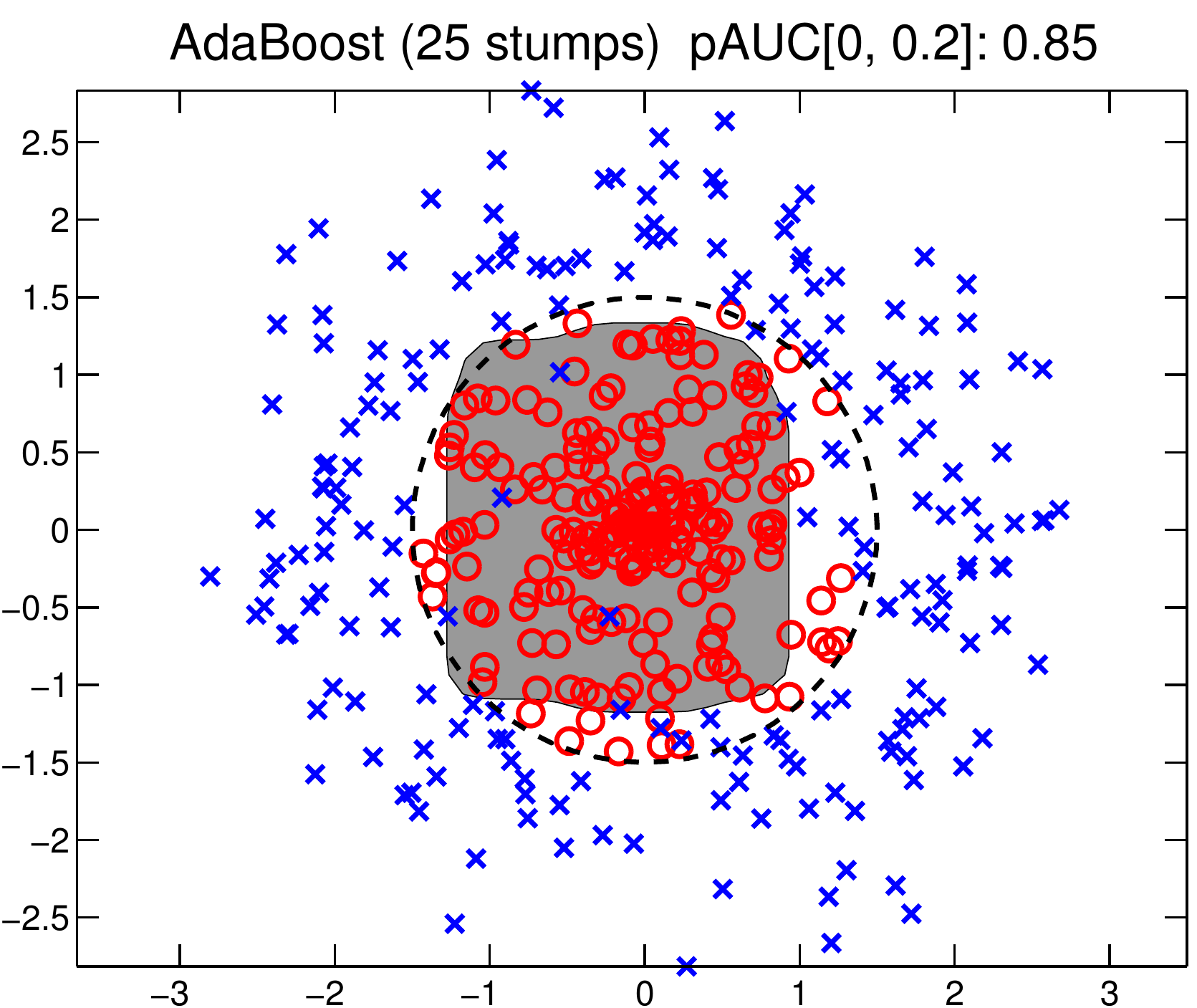}
        \includegraphics[width=0.23\textwidth,clip]{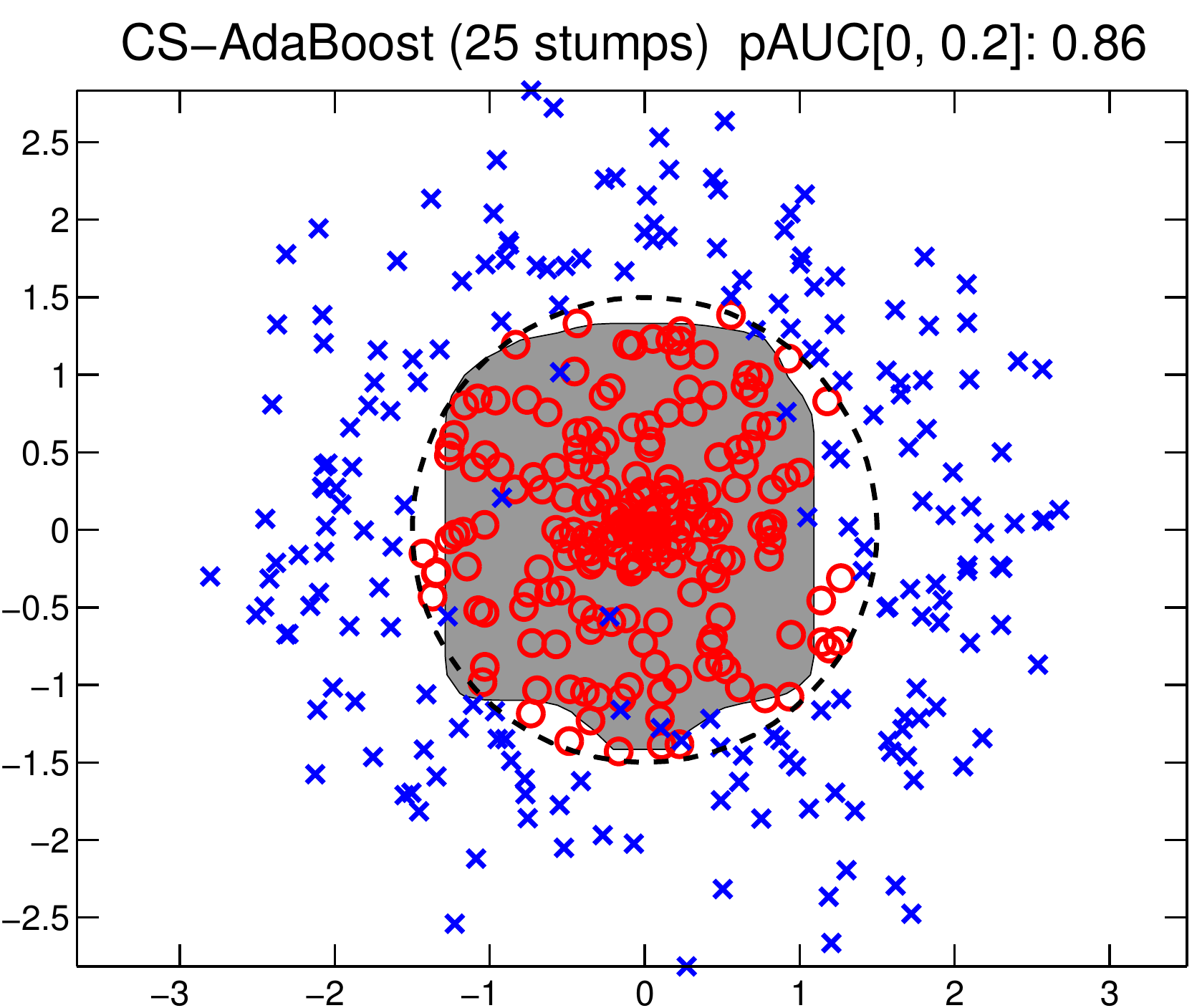}
        \includegraphics[width=0.23\textwidth,clip]{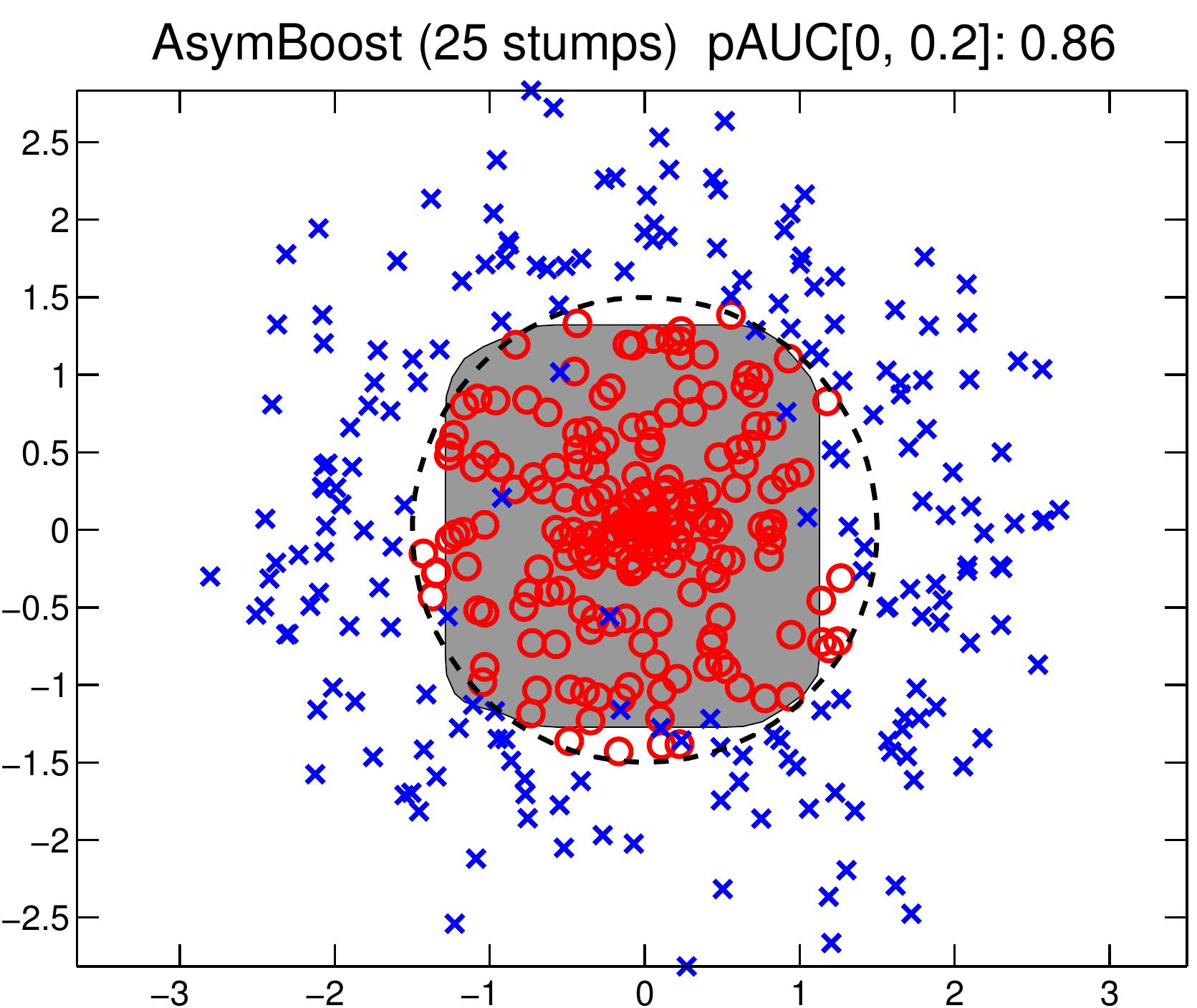}
        \includegraphics[width=0.23\textwidth,clip]{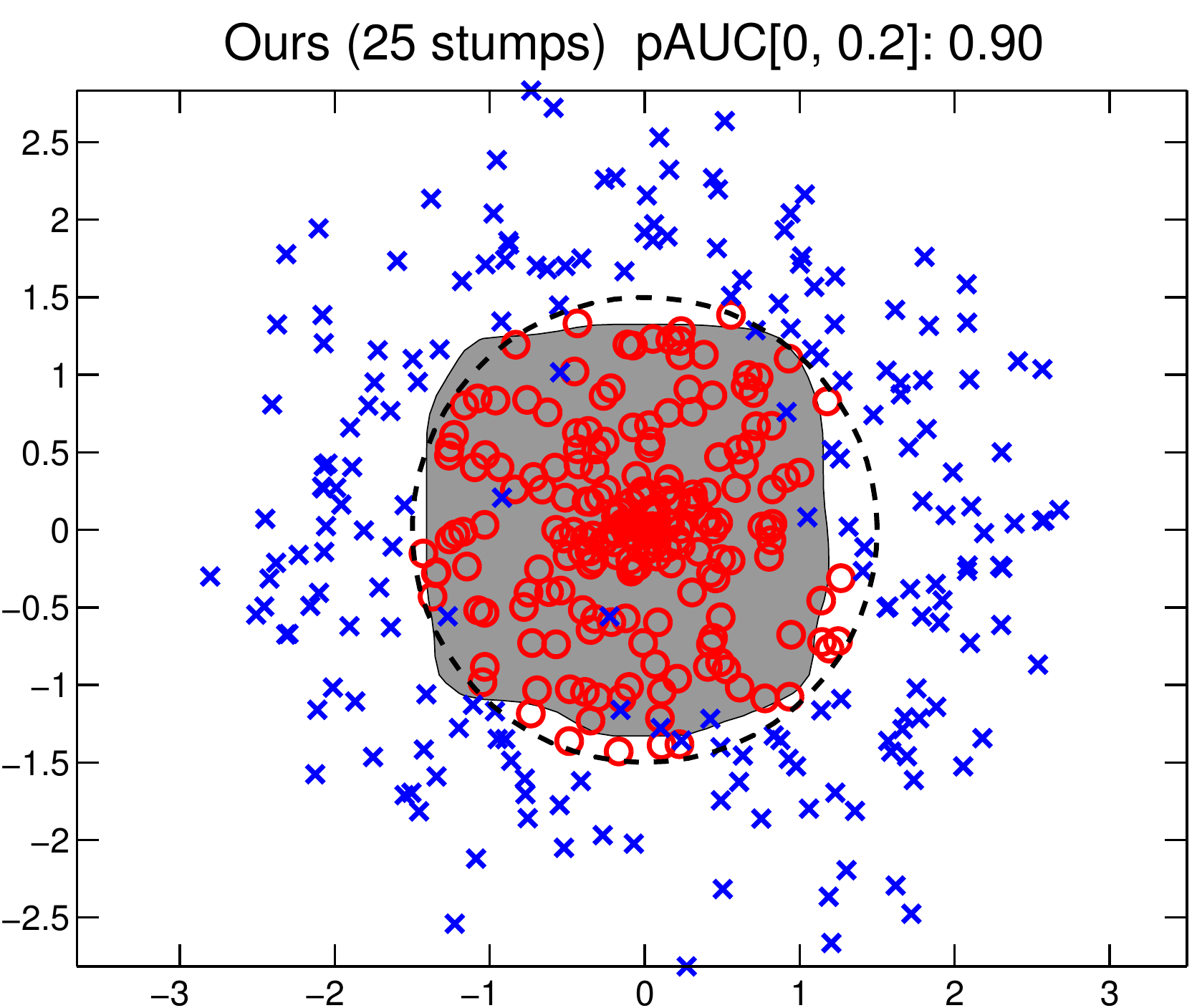}
    \caption{
    Decision boundaries on the toy data set where each strong classifier consists of
    \textbf{Top row:} $10$ weak classifiers and
    \textbf{Bottom row:} $25$ weak classifiers.
    Positive and negative data are represented by $\circ$ and $\times$, respectively.
    The partial AUC score in the \FPR range $[0, 0.2]$ is also displayed.
    Our approach achieves the best pAUC score of $0.84$ and $0.9$
    at $10$ and $25$ weak classifiers, respectively.
    At $25$ weak classifiers, we observe that both traditional and asymmetric classifiers
    start to perform similarly.
    }
    \label{fig:toy1}
  \end{figure*}

 \begin{figure*}[t]
        \centering
        \includegraphics[width=0.23\textwidth,clip]{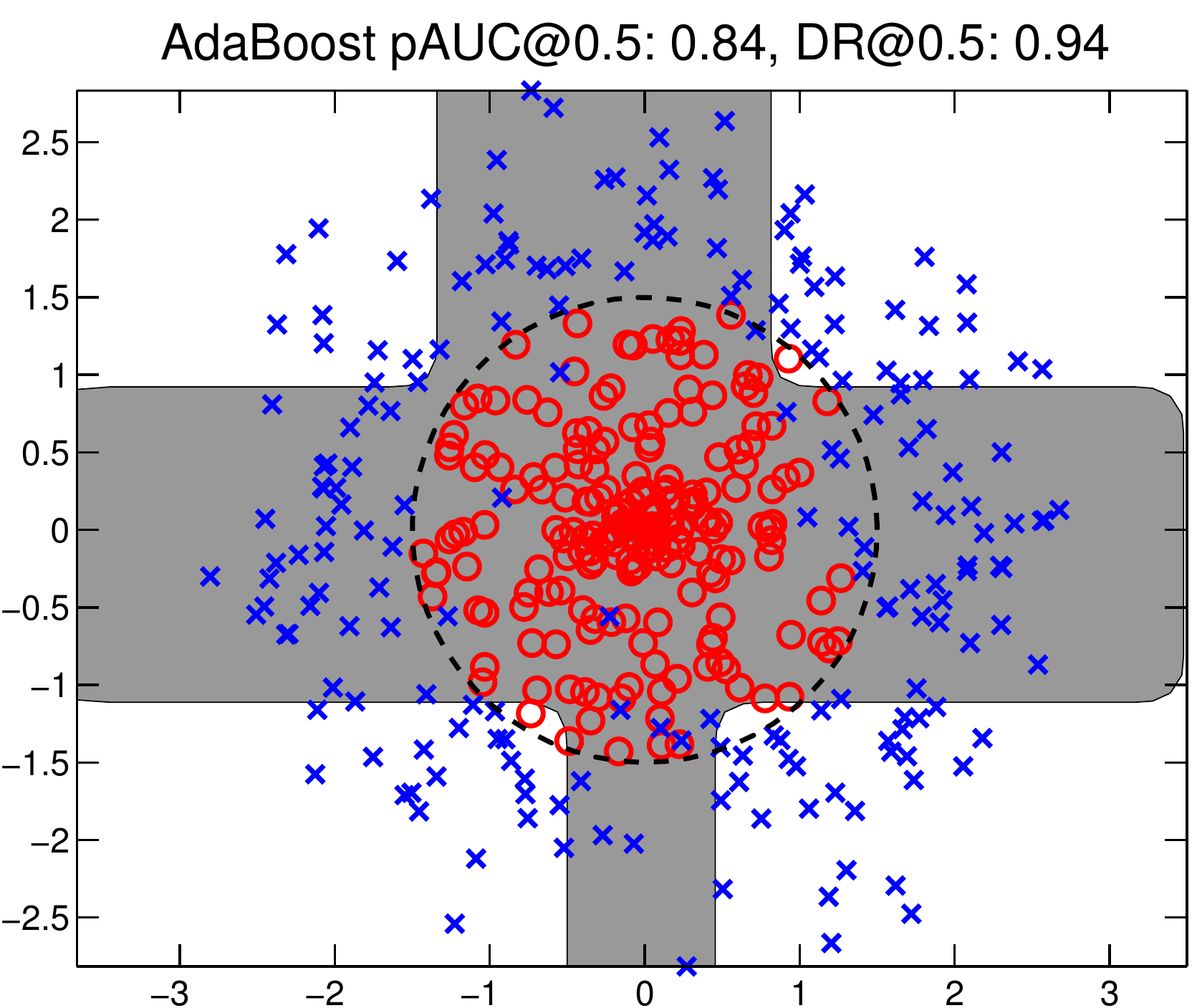}
        \includegraphics[width=0.23\textwidth,clip]{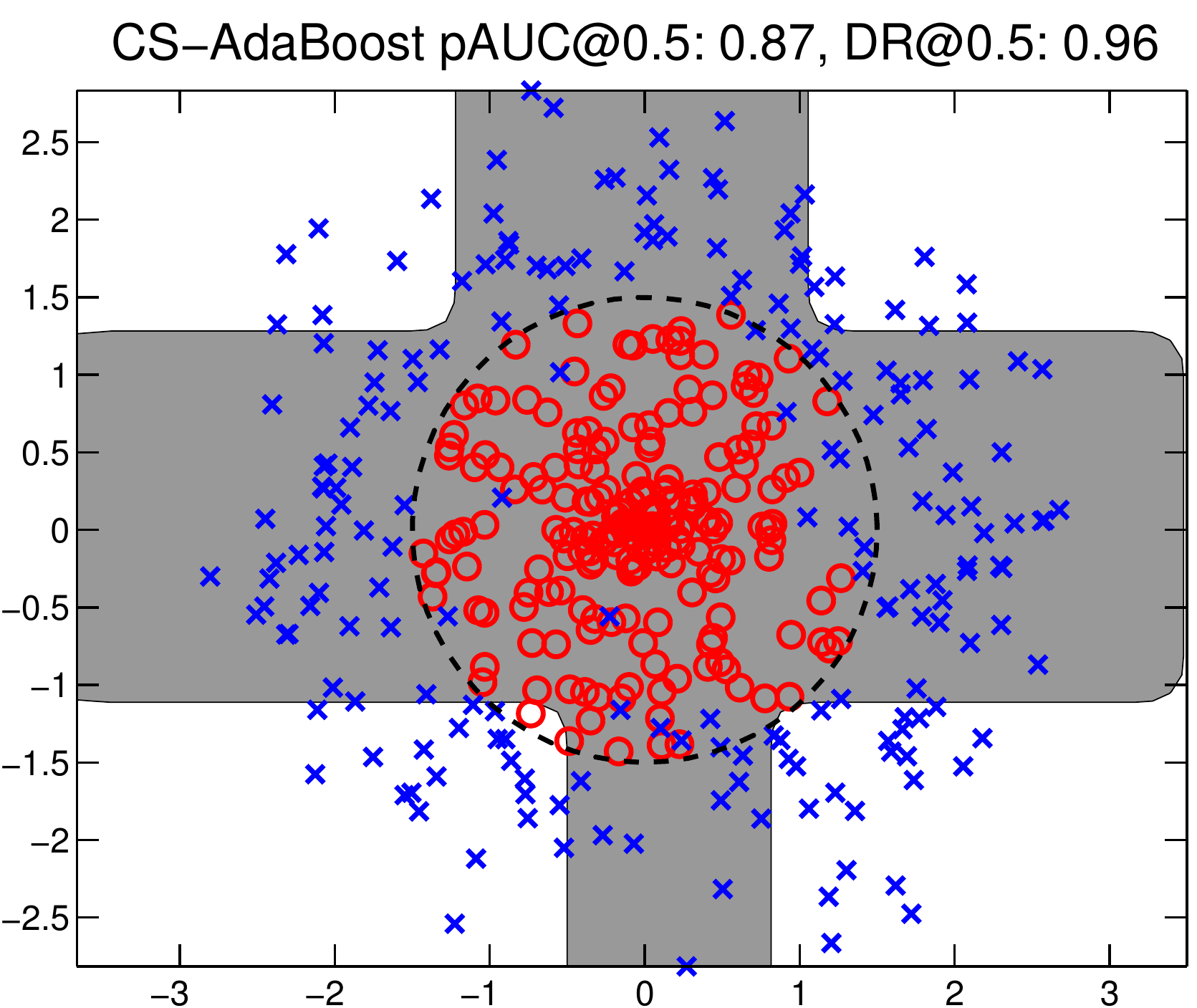}
        \includegraphics[width=0.23\textwidth,clip]{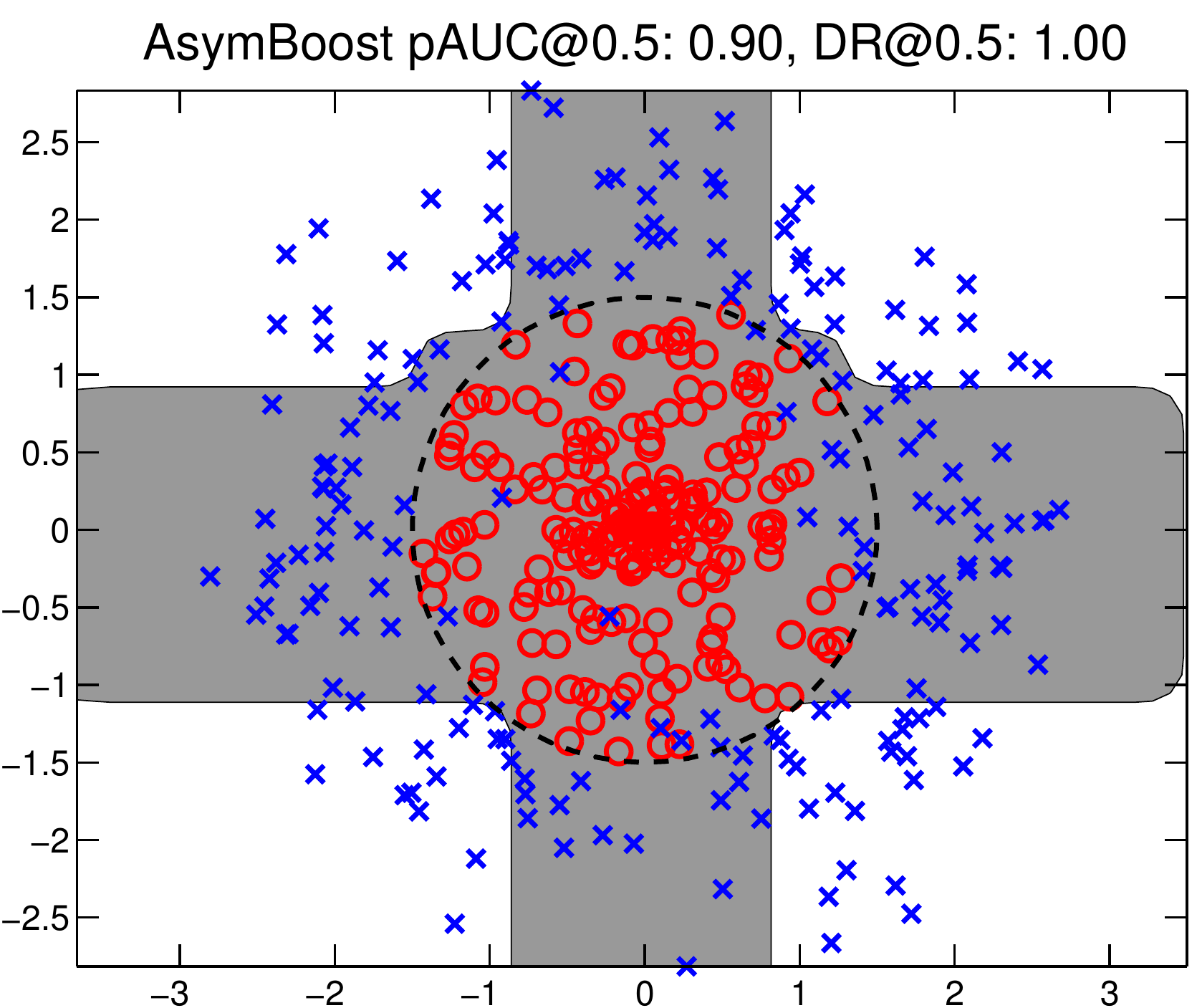}
        \includegraphics[width=0.23\textwidth,clip]{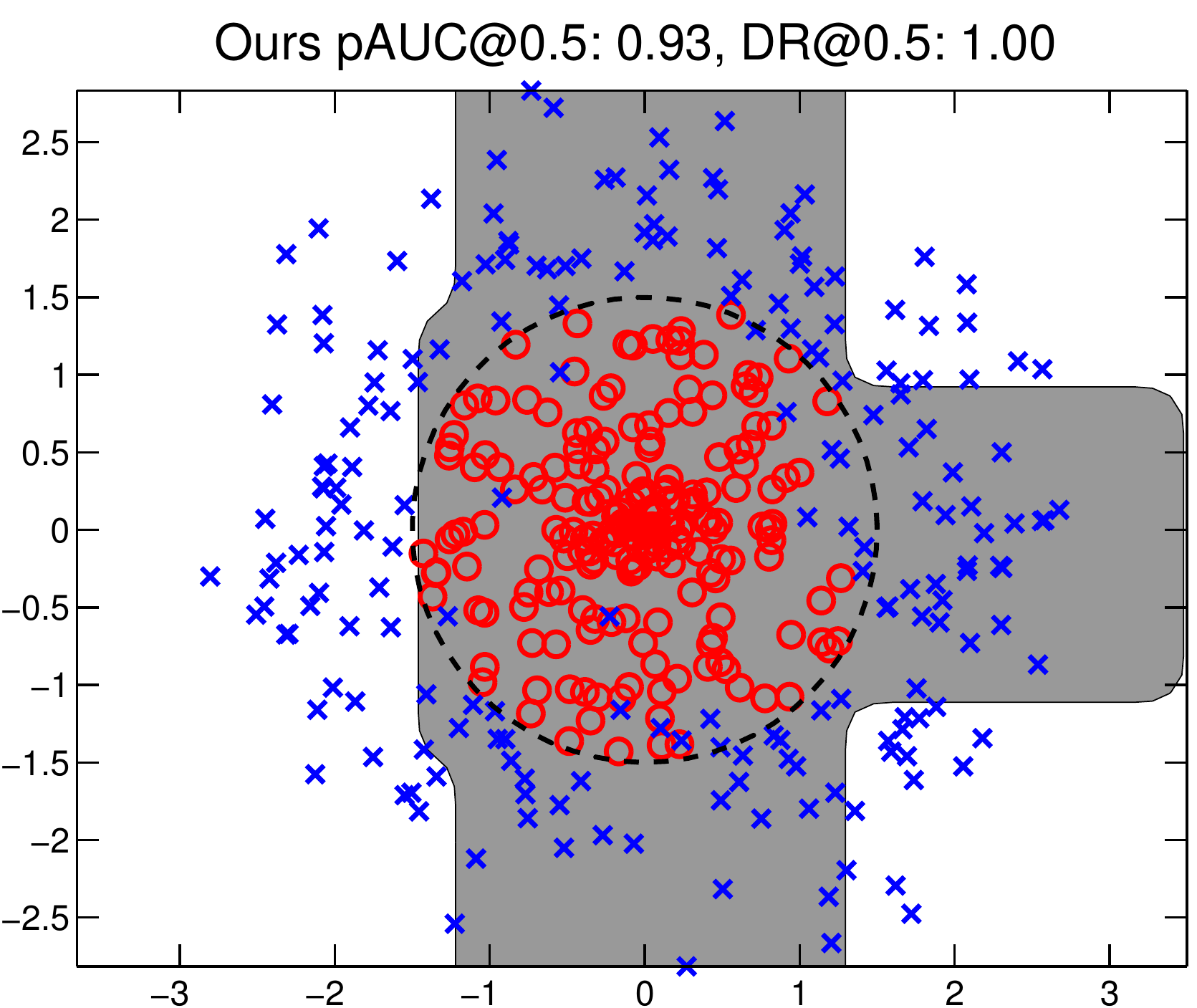}
    \caption{
    Decision boundaries on a toy data set with $10$ weak classifiers
    at \FPR of $0.5$
    The partial AUC score and detection rate at $50\%$ false positive
    rate are also shown.
    Our approach performs best on both evaluation criteria.
    Our approach preserves a larger decision boundary near
    positive samples at $\pi/4$, $3\pi/4$,
    $-\pi/4$ and $-3\pi/4$ angles.
    }
    \label{fig:toy2}
  \end{figure*}

\paragraph{Synthetic data set}
We first illustrate the effectiveness of our approach on a synthetic data set
similar to the one used in \cite{Viola2002Fast}.
We compare \algoname against the baseline AdaBoost,
Cost-Sensitive AdaBoost (CS-AdaBoost) \cite{Masnadi2011Cost} and
Asymmetric AdaBoost (AsymBoost) \cite{Viola2002Fast}.
We use vertical and horizontal decision stumps as the weak classifier.
We evaluate the partial AUC score of different algorithms at $[0, 0.2]$ FPRs.
For each algorithm, we train a strong classifier consisting of $10$ and
$25$ weak classifiers.
Additional details of the experimental set-up are provided
in the supplementary.
Fig.~\ref{fig:toy1} illustrates the boundary
decision\footnote{We set the threshold such that
the false positive rate is $0.2$.}
and the pAUC score.
Our approach outperforms all other asymmetric classifiers.
We observe that \algoname places more emphasis on
positive samples than negative samples to ensure
the highest detection rate at the left-most part of the
ROC curve (FPR $< 0.2$).
Even though we choose the asymmetric parameter, $k$, from a large range of
values, both CS-AdaBoost and AsymBoost perform slightly worse than our
approach.
AdaBoost performs worst on this toy data set
since it optimizes the overall classification accuracy.
However as the number of weak classifiers increases ($> 50$ stumps),
we observe all algorithms perform similarly on this simple toy data set.
This observation could explain the success of AdaBoost
in many object detection applications even though AdaBoost only minimizes
the symmetric error rate.

In the next experiment, we train a strong classifier of $10$ weak
classifiers and compare the performance of different classifiers
at \FPR of $0.5$.
We choose this value since it is the node learning goal often used
in training a cascade classifier.
Also we only learn $10$ weak classifiers since the first node of the
cascade often contains a small number of weak classifiers for
real-time performance.
For \algoname, we set the value of $[\alpha, \beta]$ to be $[0.49, 0.51]$.
In Fig.~\ref{fig:toy2}, we display the decision boundary of each algorithm,
and display both their pAUC score (in the \FPR range $[0.49, 0.51]$) and detection
rate at $50\%$ false positive rate.
We observe that our approach and AsymBoost have
the highest detection rate at $50\%$ false positive rate.
However, our approach outperforms AsymBoost on a pAUC score.
We observe that our approach places more emphasis on positive
samples near the corners (at $\pi/4$, $3\pi/4$,
$-\pi/4$ and $-3\pi/4$ angles) than other algorithms.

\begin{table}[bt]
  \centering
  \scalebox{1}
  {
  \begin{tabular}{l|c}
  \hline
    & pAUC($0$, $0.1$)  \\
  \hline
  \hline
   Ours (\algoname) & $\mathbf{56.05\%}$ \\
   SVM$_{\rm pAUC}$ $[0,0.1]^{\dag}$ \cite{Narasimhan2013Structural} & $54.98\%$ \\
   pAUCBoost $[0,0.1]^{\dag\dag}$ \cite{Komori2010Boosting} & $48.65\%$ \\
   Asym SVM $[0,0.1]^{\dag\dag}$ \cite{Wu2008Asymmetric} & $44.51\%$ \\
   SVM$_{\rm AUC}^{\dag\dag}$ \cite{Joachims2009Cutting} & $39.72\%$ \\
  \hline
  \end{tabular}
  }
  \caption{The pAUC score on Protein-protein interaction data set.
  The \textbf{higher the pAUC score}, the \textbf{better the classifier}.
  $\dag$ The result reported here is better than the one reported in
  \cite{Narasimhan2013Structural}.
  We suspect that we tuned the regularization parameter in the finer range.
  Results marked by $\dag\dag$ were reported in \cite{Narasimhan2013Structural}.
  The best classifier is shown in boldface.
  }
  \label{tab:ppi}
\end{table}

\paragraph{Protein-protein interaction prediction}
In this experiment, we compare our approach with
existing algorithms which optimize \pAUC in bioinformatics.
The problem we consider here is a protein-protein interaction
prediction \cite{Qi2006Evaluation}, in which the task is to predict
whether a pair of proteins interact or not.
We used the data set labelled `Physical Interaction Task
in Detailed feature type', which is publicly available
on the internet\footnote{\url{http://www.cs.cmu.edu/~qyj/papers_sulp/proteins05_pages/feature-download.html}}.
The data set contains $2865$ protein pairs known to be interacting (positive)
and a random set of $237,384$ protein pairs labelled as non-interacting (negative).
We use a subset of $85$ features as in \cite{Narasimhan2013Structural}.
We randomly split the data into two groups: $10\%$ for training/validation
and $90\%$ for evaluation.
We choose the best regularization parameter form $\{5$, $2$, $1$, $1/2$,
$1/5\}$ by $5$-fold cross validation.
We repeat our experiments $10$ times using the same regularization parameter.
We train a linear classifier as our weak learner using
\rm{LIBLINEAR} \cite{Fan2008Liblinear}.
We set the maximum number of boosting iterations to $100$
and report the pAUC score of our approach in Table~\ref{tab:ppi}.
Baselines include SVM$_{\rm pAUC }$,
SVM$_{\rm AUC }$, pAUCBoost and Asymmetric SVM.
Our approach outperforms all existing algorithms
which optimize either \AUC or \pAUC.
We attribute our improvement over
SVM$_{\rm pAUC}$ $[0,0.1]$ \cite{Narasimhan2013Structural}, as a result of introducing
a non-linearity into the original problem.
This phenomenon has also been observed in face detection
as reported in \cite{Wu2008Fast}.

\paragraph{Comparison to other asymmetric boosting}
Here we compare \algoname against
several boosting algorithms previously proposed for the problem of
object detection, namely,
AdaBoost with Fisher LDA post-processing \cite{Wu2008Fast},
AsymBoost \cite{Viola2002Fast} and CS-AdaBoost \cite{Masnadi2011Cost}.
The results of AdaBoost are also presented as the baseline.
For each algorithm, we train a strong classifier consisting of
$100$ weak classifiers.
We then calculate the pAUC score by varying the threshold value
in the \FPR range $[0, 0.1]$.
For each algorithm, the experiment is repeated $20$ times and
the average pAUC score is reported.
For AsymBoost, we choose $k$ from  $\{2^{-0.5}$, $2^{-0.4}$, $\cdots$,
$2^{0.5} \}$ by cross-validation.
For CS-AdaBoost, we choose $k$ from $\{0.5$, $0.75$, $\cdots$, $3\}$
by cross-validation.
We evaluate the performance of all algorithms on $3$
vision data sets: USPS digits, scenes and face data sets.
See supplementary for more details on feature extraction.
We report the experimental results in Table~\ref{tab:expvision}.
From the table, \algoname demonstrates the best performance on all
three vision data sets.

\begin{table}[t]
  \centering
  \scalebox{0.8}
  {
  \begin{tabular}{c|c|ccc}
  \hline
 &  $\#$ iters & USPS  & SCENE  & FACE  \\
\hline
\hline
  Ours & $10$ & $\mathbf{0.77}$ ($\mathbf{0.02}$)  & $\mathbf{0.65}$ ($\mathbf{0.02}$)  & $\mathbf{0.53}$ ($\mathbf{0.02}$) \\
  (\algoname) & $20$ & $\mathbf{0.85}$ ($\mathbf{0.01}$)  & $\mathbf{0.75}$ ($\mathbf{0.01}$)  & $\mathbf{0.67}$ ($\mathbf{0.01}$) \\
   & $100$ & $\mathbf{0.93}$ ($\mathbf{0.01}$)  & $\mathbf{0.85}$ ($\mathbf{0.01}$)  & $\mathbf{0.82}$ ($\mathbf{0.01}$) \\
\hline
  AdaBoost  & $10$ & $0.75$ ($0.03$)  & $0.63$ ($0.03$)  & $0.52$ ($0.03$) \\
  \cite{Viola2004Robust} & $20$ & $0.82$ ($0.03$)  & $0.72$ ($0.03$)  & $0.63$ ($0.03$) \\
   & $100$ & $0.90$ ($0.02$)  & $0.83$ ($0.02$)  & $0.78$ ($0.02$) \\
\hline
  Ada + LDA & $10$ & $0.74$ ($0.02$)  & $0.63$ ($0.02$)  & $\mathbf{0.53}$ ($\mathbf{0.02}$) \\
  \cite{Wu2008Fast} & $20$ & $0.79$ ($0.02$)  & $0.73$ ($0.02$)  & $0.63$ ($0.02$) \\
   & $100$ & $0.86$ ($0.01$)  & $0.84$ ($0.01$)  & $0.76$ ($0.01$) \\
\hline
  AsymBoost & $10$ & $0.76$ ($0.02$)  & $0.58$ ($0.02$)  & $0.51$ ($0.02$) \\
  \cite{Viola2002Fast} & $20$ & $0.83$ ($0.01$)  & $0.69$ ($0.01$)  & $0.65$ ($0.01$) \\
   & $100$ & $0.85$ ($0.03$)  & $0.81$ ($0.03$)  & $0.82$ ($0.03$) \\
\hline
  CS-AdaBoost  & $10$ & $0.75$ ($0.04$)  & $0.64$ ($0.04$)  & $0.52$ ($0.04$) \\
  \cite{Masnadi2011Cost} & $20$ & $0.82$ ($0.03$)  & $0.73$ ($0.03$)  & $0.63$ ($0.03$) \\
   & $100$ & $0.91$ ($0.02$)  & $0.84$ ($0.02$)  & $0.78$ ($0.02$) \\
\hline
  \end{tabular}
  }
  \caption{Average pAUC scores and their standard deviations
  on vision data sets at various boosting iterations.
  All experiments are repeated $20$ times.
  The best average performance is shown in boldface.
  }
  \label{tab:expvision}
\end{table}

\paragraph{Pedestrian detection - Strong classifier}
We evaluate our approach on the pedestrian detection task.
We train our approach on the INRIA pedestrian data set.
For the positive training data,
we use all $2416$ INRIA cropped pedestrian images.
To generate the negative training data,
we first train the cascade classifier with $20$ nodes using
Viola and Jones' approach.
We then combine $2416$ random negative windows generated
in the first node with another $4832$ negative windows generated
in the subsequent nodes.
The resulting $7248$ negative windows are used for training
the strong classifier.
We generate a large pool of features by combining
the histogram of oriented gradient (HOG) features \cite{Dalal2005HOG}
and covariance (COV)
features\footnote{Covariance features capture the
relationship between different image statistics and have been
shown to perform well in our previous experiments.
However, other discriminative features can also be used here instead,
\eg, Haar-like features,
Local Binary Pattern (LBP) \cite{mu2008lbp},
Sketch Tokens \cite{Lim2013Sketch} and
self-similarity of low-level features (CSS) \cite{Walk2010New}.}
\cite{Tuzel2008Pedestrian}.
Additional details of HOG and COV parameters
are provided in the supplementary.
We use weighted linear discriminant analysis (WLDA) as weak classifiers.
We train $500$ weak classifiers and set $5$ multi-exits \cite{Pham2008Detection}.
To be more specific, we set the threshold at
$10$, $20$, $50$, $100$ and $200$ weak classifiers.
These exits reduce the evaluation time during testing significantly.
The regularization parameter $\nu$ is cross-validated from $\{0.1, 0.5, 1, 2, 10\}$.
Since we have not carefully cross-validated a finer range of $\nu$,
tuning this parameter could yield a further improvement.
The training time of our approach is under two hours on a parallelized quad core Xeon machine.

During evaluation, each test image is scanned with $4 \times 4$ pixels step
stride and the scale ratio of input image pyramid is $1.05$.
The overlapped detection windows are merged using the greedy
non-maximum suppression strategy as introduced
in \cite{Dollar2009Integral}.
We use the continuous AUC evaluation software of
Sermanet \etal \cite{Sermanet2013Pedestrian} and
report the pAUC score between $[0, 0.005]$ \FPPI ($1$ false positive),
$[0, 0.05]$ \FPPI ($15$ false positives),
$[0, 0.5]$ \FPPI ($144$ false positives) and
$[0, 1]$ \FPPI ($288$ false positives) in Table~\ref{tab:strongbeta}.
From the table, we observe that setting the value of $\beta$
to be minimal ($\beta = 0.05$) yeilds the best pAUC score at $[0, 0.005]$ \FPPI.
As we increase the \FPPI range, the higher value of $\beta$ tends to perform
better.
This table clearly illustrates the advantage of our approach.

\begin{table}[bt]
  \centering
  \scalebox{.9}
  {
  \begin{tabular}{l|cccc}
  \hline
    \backslashbox{Train}{Test (FPPI)}
    &  \makebox[2.5em]{$[0,0.005]$} & \makebox[2.5em]{$[0,0.05]$}
    &  \makebox[2.5em]{$[0,0.5]$}  & \makebox[2.5em]{$[0,1]$}  \\
  \hline
  \hline
   $\beta=0.05$ & $\mathbf{65.0\%}$ & $39.3\%$ & $20.6\%$ & $18.1\%$ \\
   $\beta=0.1$ & $75.7\%$ & $\mathbf{32.5\%}$ & $18.0\%$ & $15.8\%$ \\
   $\beta=0.5$ & $73.4\%$ & $33.8\%$ & $\mathbf{17.9\%}$ & $\mathbf{15.3\%}$ \\
  \hline
  \end{tabular}
  }
  \caption{The pAUC score in the \FPR range $[0, \beta]$
  on the training set.
  Our objective here is to optimize the area under the curve between
  $[0,0.005]$, $[0,0.05]$, $[0,0.5]$ and $[0,1]$ \FPPI on
  the INRIA test set.
  Since we plot \FPPI versus
  miss rate, a \textbf{smaller pAUC score} means a \textbf{better detector}.
  The best detector at each \FPPI range is shown in boldface.
  Clearly, a large value of $\beta$ is best for
  a large \FPPI range.
  }
  \label{tab:strongbeta}
\end{table}

\begin{figure}[t]
        \centering
        \includegraphics[width=0.48\textwidth,clip]{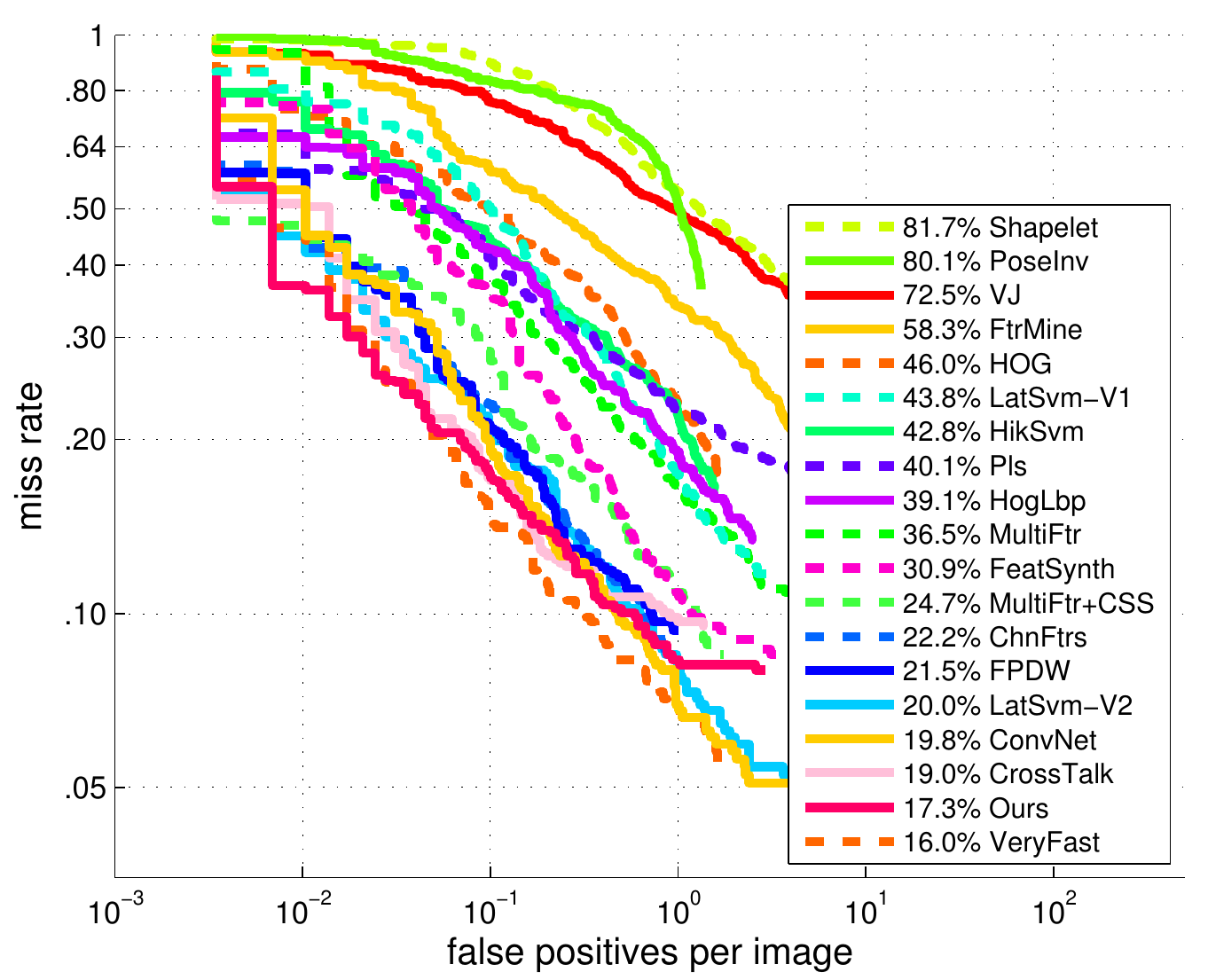}
    \caption{
    ROC curves of our approach and several state-of-the-art
    detectors on the INRIA test image.
    We train a strong classifier using HOG and COV features.
    }
    \label{fig:strong1}
\end{figure}

Fig.~\ref{fig:strong1} compares the performance of
our approach with other state-of-the-art algorithms
on the INRIA pedestrian data set.
We use the evaluation software of Doll\'{a}r \etal \cite{Dollar2012Pedestrian},
which computes
the AUC from $9$ discrete points sampled between $[0.01, 1.0]$ \FPPI.
Our approach performs second best on this data set.
It performs comparable to VeryFast \cite{Benenson2012Pedestrian}
which trains multiple detectors at multiple scale.
Upon a closer observation, our \algoname performs slightly
better than VeryFast when the number of \FPPI
is less then $0.1$ and VeryFast performs slightly better
when the number of \FPPI
is greater $0.1$.
We evaluate our strong classifier on TUD-Brussels and ETH
pedestrian data sets but we observe that the detection
results contain a large number of false positives.
Instead of bootstrapping with more negative samples
as in \cite{Dollar2009Integral, Walk2010New},
we train a cascade classifier in the next section.

 \begin{figure}[t]
        \centering
        \includegraphics[width=0.4132\textwidth,clip]{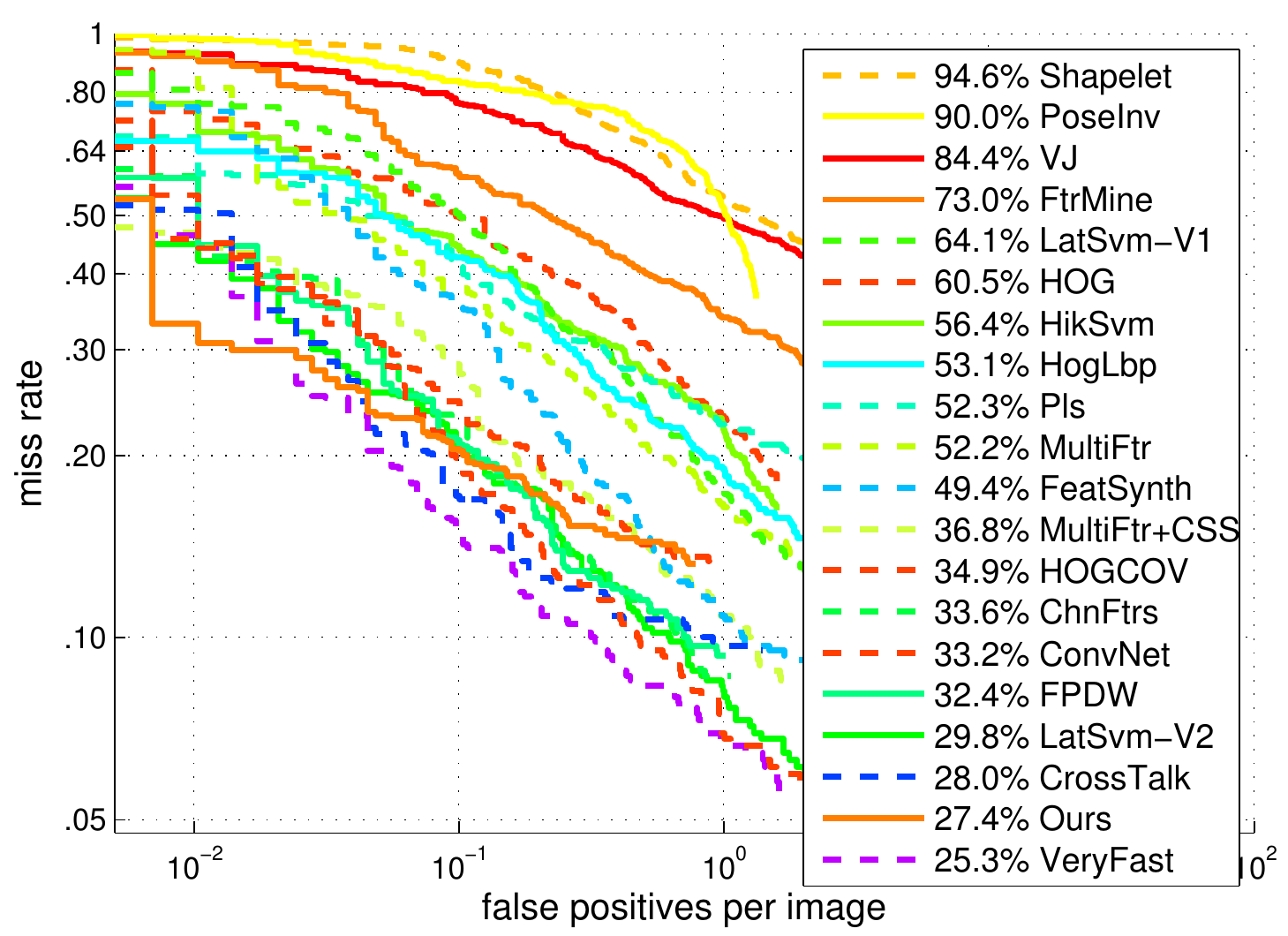}
        \includegraphics[width=0.4132\textwidth,clip]{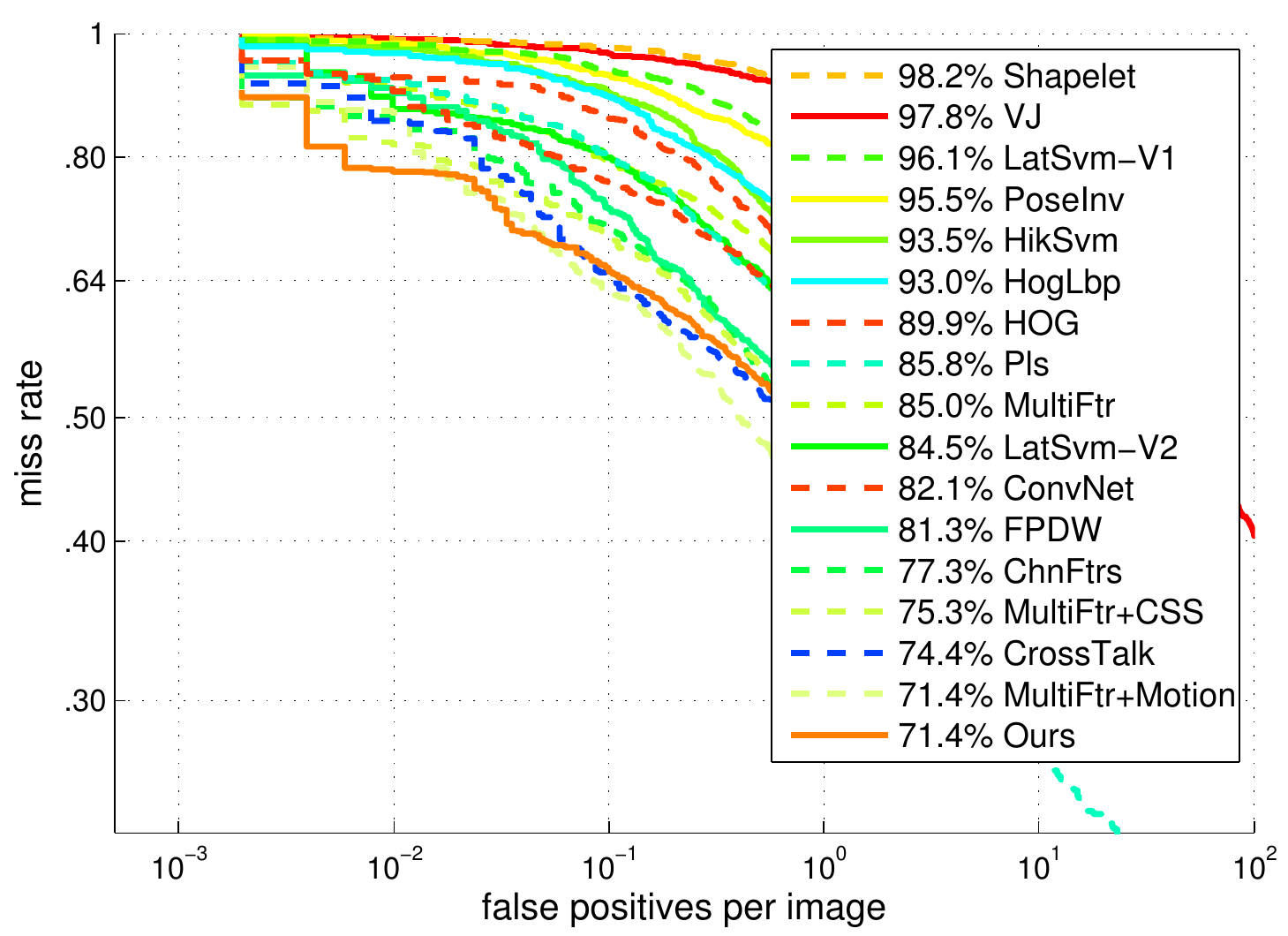}
        \includegraphics[width=0.4132\textwidth,clip]{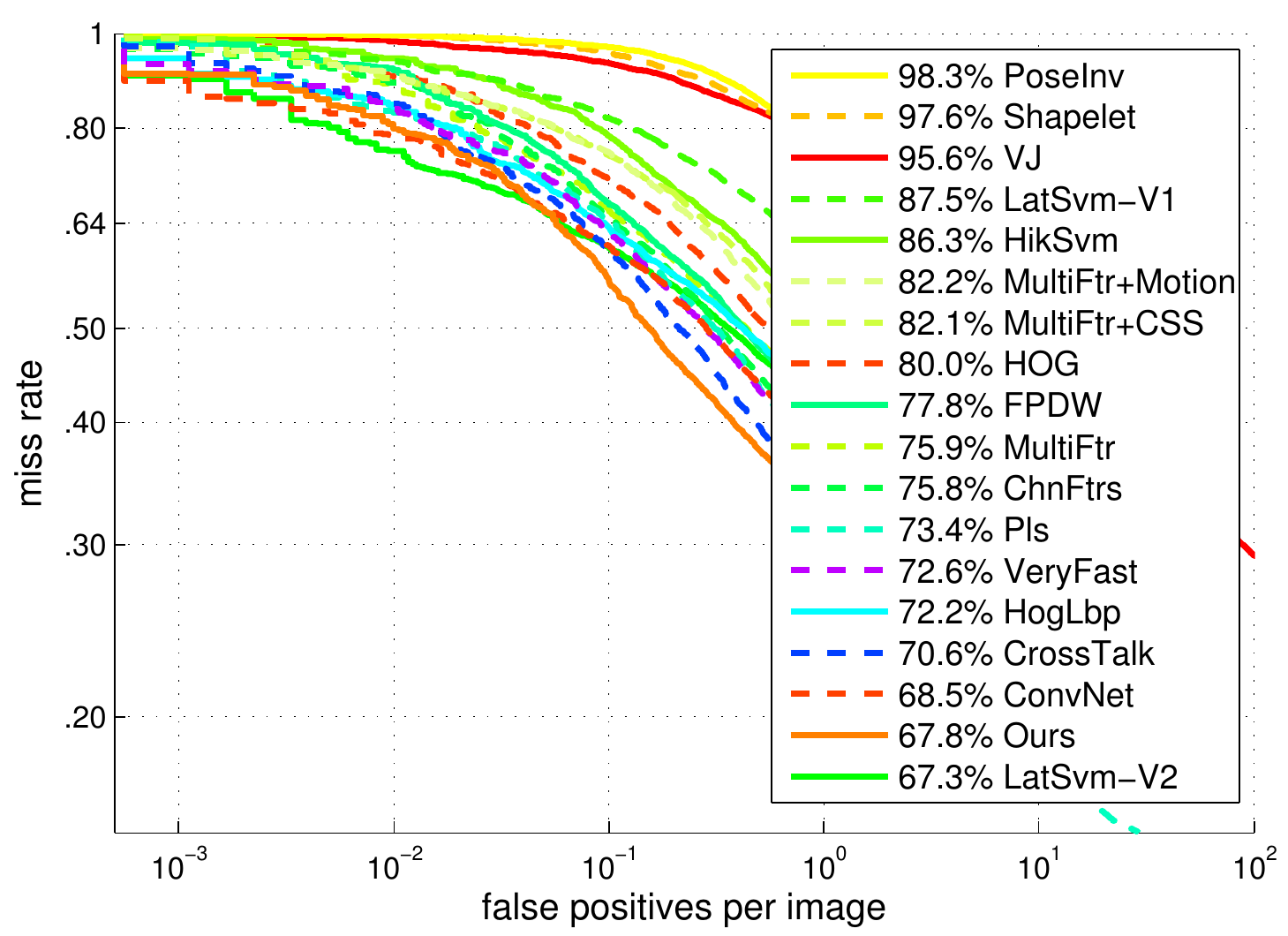}
    \caption{
    From top to bottom: performance on INRIA, TUD-Brussels and ETH test set.
    Algorithms are sorted using the \textbf{partial AUC score} in the \FPPI range $[0, 0.1]$.
    Our \algoname consistently performs comparable to the state-of-the-art.
    }
    \label{fig:cascade}
  \end{figure}

\paragraph{Pedestrian detection - Cascade classifier}
In this section, we train a cascade classifier using our \algoname.
We train our detector on INRIA training set and evaluate the detector on
INRIA, TUD-Brussels and ETH test sets.
On both TUD-Brussels and ETH data sets, we upsample
the original image to $1600 \times 1200$ pixels before
applying our pedestrian detector.
We train the human detector with a combination of
HOG and COV features as previously described.
To achieve the node learning goal of the cascade
(each node achieves an extremely high detection rate ($> 99\%$)
and a moderate false positive rage ($\approx 50\%$)),
we optimize the pAUC in the \FPR range $[0.49, 0.51]$.
We train a multi-exit cascade \cite{Pham2008Detection}
with $19$ exit.
In this experiment, we use the software of \cite{Sermanet2013Pedestrian}
to compute the continuous AUC score
in the \FPPI range $[0, 0.1]$.
We sort different algorithms
based on the pAUC score in the \FPPI range $[0, 0.1]$
and report the results in Fig.~\ref{fig:cascade}.
We compare our proposed approach with the
baseline HOGCOV classifier (using AdaBoost).
We observe that our approach reduces the average miss-rate over
HOGCOV by $7\%$ on INRIA test set.
From Fig.~\ref{fig:cascade}, our approach achieves similar performance to the state-of-the-art
detector.
We then break-down experimental results of different measures
using the partial AUC score
(\FPPI range $[0,0.1]$) in Table~\ref{tab:cascade}.
On average, our approach performs best on the {\em large} evaluation setting
where pedestrians are at least $100$ pixels tall.
On other settings, our approach yields competitive results to the
state-of-the-art detector in that category.
In summary, our approach performs better than or on par with
the state-of-the-art despite its simplicity
(in comparison to LatSvm --- a part-based approach
which models unknown parts as latent variables).
In addition, the current detector is only trained with
two discriminative visual features (HOG and COV).
Applying additional discriminative features, \eg,
LBP \cite{Wang2009HOG} or motion features \cite{Walk2010New},
could further improve the overall detection performance.

\begin{table*}[htb]
\MyTiny{
\setlength{\tabcolsep}{.3em}
\begin{tabular}{ c | c | c c c c c c c c c c c c c c c c c c c c }
\hline
& \textbf{Ours}& \textbf{ChnFtrs}& \textbf{ConvNet}& \textbf{CrossTalk}& \textbf{FPDW}& \textbf{FeatSynth}& \textbf{FtrMine}& \textbf{HOG}& \textbf{HikSvm}& \textbf{HogLbp}& \textbf{LatSvm-V1}& \textbf{LatSvm-V2}& \textbf{MultiFtr}& \textbf{Pls}& \textbf{PoseInv}& \textbf{Shapelet}& \textbf{VJ}& \textbf{VeryFast}\\
\hline
%
\hline
\multicolumn{19}{c}{Reasonable (min. 50 pixels tall \& no/partial occlusion) - Partial AUC(0,0.1)\%}\\
\hline
INRIA-Fixed & 27.4 & 31.6 & 31.9 & 26.9 & 30.9 & 49.3 & 71.5 & 59.4 & 53.9 & 50.5 & 62.6 & 29.1 & 51.3 & 50.4 & 89.4 & 93.0 & 83.2 & \textbf{23.9}\\
TudBrussels & \textbf{65.8} & 72.2 & 77.8 & 68.8 & 77.0 & - & - & 87.9 & 92.3 & 91.3 & 95.5 & 80.8 & 81.6 & 82.6 & 94.5 & 97.8 & 97.4 & -\\
ETH & 62.8 & 72.4 & 62.8 & 67.0 & 74.5 & - & - & 78.1 & 85.5 & 67.4 & 86.1 & \textbf{61.4} & 73.1 & 69.5 & 98.1 & 97.3 & 95.4 & 68.7\\
\hline
\multicolumn{19}{c}{Large (min. 100 pixels tall) - Partial AUC(0,0.1)\%}\\
\hline
INRIA-Fixed & 26.0 & 29.6 & 27.6 & 25.0 & 28.7 & 48.6 & 70.9 & 59.0 & 53.1 & 49.5 & 61.6 & 25.9 & 50.0 & 49.3 & 89.4 & 92.9 & 82.9 & \textbf{21.7}\\
TudBrussels & \textbf{47.3} & 50.0 & 49.5 & 52.8 & 53.8 & - & - & 88.4 & 85.1 & 73.9 & 88.9 & 67.9 & 71.4 & 66.8 & 94.2 & 92.8 & 96.3 & -\\
ETH & \textbf{42.9} & 57.6 & 48.1 & 48.6 & 62.7 & - & - & 56.9 & 66.0 & 54.2 & 74.7 & 45.5 & 59.4 & 50.8 & 96.3 & 93.4 & 92.0 & 48.6\\
\hline
\multicolumn{19}{c}{Near (min. 80 pixels tall) - Partial AUC(0,0.1)\%}\\
\hline
INRIA-Fixed & 25.9 & 30.1 & 30.8 & 25.4 & 29.5 & 48.3 & 70.9 & 58.5 & 53.0 & 49.4 & 62.0 & 27.6 & 50.3 & 49.5 & 89.2 & 92.9 & 82.8 & \textbf{22.7}\\
TudBrussels & \textbf{49.0} & 59.1 & 60.1 & 58.7 & 62.1 & - & - & 87.1 & 88.0 & 79.5 & 91.7 & 70.6 & 74.5 & 75.1 & 93.9 & 95.1 & 96.3 & -\\
ETH & 51.3 & 64.8 & 51.8 & 55.9 & 66.5 & - & - & 69.2 & 74.2 & 56.7 & 77.5 & \textbf{51.0} & 63.2 & 60.9 & 97.8 & 95.2 & 93.6 & 55.4\\
\hline
\multicolumn{19}{c}{Medium (min. 30 pixels tall and max. 80 pixels tall) - Partial AUC(0,0.1)\%}\\
\hline
INRIA-Fixed & 100.0 & 100.0 & 54.9 & 96.5 & 100.0 & 100.0 & 100.0 & 100.0 & 100.0 & 94.6 & 88.8 & 96.5 & 94.3 & 100.0 & 96.5 & 96.5 & 89.6 & \textbf{51.3}\\
TudBrussels & 75.3 & 78.0 & 84.6 & \textbf{74.0} & 81.3 & - & - & 86.4 & 93.3 & 97.0 & 96.1 & 85.7 & 83.7 & 86.3 & 94.0 & 98.6 & 97.3 & -\\
ETH & 67.2 & \textbf{64.8} & 76.0 & 65.0 & 67.1 & - & - & 69.8 & 80.9 & 78.7 & 88.3 & 76.7 & 68.6 & 67.0 & 96.3 & 89.8 & 89.0 & 73.2\\
\hline
\end{tabular}
}
  \caption{Performance comparison of various detectors on several pedestrian test sets.
The best detector in each category from each data set is highlighted in
bold.
The \AUC score is taken over the \FPPI range $[0, 0.1]$.
A \textbf{smaller pAUC score} means a \textbf{better detector}.
The \AUC score over the \FPPI range $[0, 1]$
can be found in the supplementary.
  }
  \label{tab:cascade}
\end{table*}

\section{Conclusion}
We have proposed a new ensemble learning method
for object detection.
The proposed approach is based on optimizing
the partial \AUC score in the \FPR range $[\alpha, \beta]$.
Extensive experiments demonstrate the effectiveness of the
proposed approach in visual detection tasks.
We plan to explore the possibility of applying the
proposed approach to the multiple scales detector of \cite{Benenson2012Pedestrian}
in order to improve the detection results of very low
resolution pedestrian images.

{\small
\bibliographystyle{abbrv}
\bibliography{draft}

\begin{thebibliography}{10}

\bibitem{Benenson2012Pedestrian}
R.~Benenson, M.~Mathias, R.~Timofte, and L.~V. Gool.
\newblock Pedestrian detection at 100 frames per second.
\newblock In {\em Proc. IEEE Conf. Comp. Vis. Patt. Recogn.}, 2012.

\bibitem{Benenson2013Seeking}
R.~Benenson, M.~Mathias, T.~Tuytelaars, and L.~V. Gool.
\newblock Seeking the strongest rigid detector.
\newblock In {\em Proc. IEEE Conf. Comp. Vis. Patt. Recogn.}, 2013.

\bibitem{Dalal2005HOG}
N.~Dalal and B.~Triggs.
\newblock Histograms of oriented gradients for human detection.
\newblock In {\em Proc. IEEE Conf. Comp. Vis. Patt. Recogn.}, volume~1, 2005.

\bibitem{demiriz2002}
A.~Demiriz, K.~Bennett, and J.~Shawe-Taylor.
\newblock Linear programming boosting via column generation.
\newblock {\em Mach. Learn.}, 46(1--3):225--254, 2002.

\bibitem{Dodd2003Partial}
L.~E. Dodd and M.~S. Pepe.
\newblock Partial auc estimation and regression.
\newblock {\em Biometrics}, 59(3):614--623, 2003.

\bibitem{Dollar2009Integral}
P.~Doll{\'a}r, Z.~Tu, P.~Perona, and S.~Belongie.
\newblock Integral channel features.
\newblock In {\em Proc. of British Mach. Vis. Conf.}, 2009.

\bibitem{Dollar2012Pedestrian}
P.~Doll\'ar, C.~Wojek, B.~Schiele, and P.~Perona.
\newblock Pedestrian detection: An evaluation of the state of the art.
\newblock {\em {IEEE} Trans. Pattern Anal. Mach. Intell.}, 34(4):743--761,
  2012.

\bibitem{Fan2008Liblinear}
R.-E. Fan, K.-W. Chang, C.-J. Hsieh, X.-R. Wang, and C.-J. Lin.
\newblock {LIBLINEAR}: A library for large linear classification.
\newblock {\em J. Mach. Learn. Res.}, 9:1871--1874, 2008.

\bibitem{Hsu2012Linear}
M.-J. Hsu and H.-M. Hsueh.
\newblock The linear combinations of biomarkers which maximize the partial area
  under the roc curves.
\newblock {\em Comp. Stats.}, 28(2):1--20, 2012.

\bibitem{Joachims2009Cutting}
T.~Joachims, T.~Finley, and C.-N.~J. Yu.
\newblock Cutting-plane training of structural svms.
\newblock {\em Mach. Learn.}, 77(1):27--59, 2009.

\bibitem{Komori2010Boosting}
O.~Komori and S.~Eguchi.
\newblock A boosting method for maximizing the partial area under the roc
  curve.
\newblock {\em BMC Bioinformatics}, 11(1):314, 2010.

\bibitem{Komori2011Boosting}
O.~Komori and S.~Eguchi.
\newblock Boosting learning algorithm for pattern recognition and beyond.
\newblock {\em IEICE Trans. Infor. and Syst.}, 94(10):1863--1869, 2011.

\bibitem{Lim2013Sketch}
J.~J. Lim, C.~L. Zitnick, and P.~Dollar.
\newblock {S}ketch {T}okens: A learned mid-level representation for contour and
  object detection.
\newblock In {\em Proc. IEEE Conf. Comp. Vis. Patt. Recogn.}, 2013.

\bibitem{Masnadi2011Cost}
H.~Masnadi-Shirazi and N.~Vasconcelos.
\newblock Cost-sensitive boosting.
\newblock {\em {IEEE} Trans. Pattern Anal. Mach. Intell.}, 33(2):294--309,
  2011.

\bibitem{mu2008lbp}
Y.~Mu, S.~Yan, Y.~Liu, T.~Huang, and B.~Zhou.
\newblock Discriminative local binary patterns for human detection in personal
  album.
\newblock In {\em Proc. IEEE Conf. Comp. Vis. Patt. Recogn.}, Anchorage, AK,
  US, 2008.

\bibitem{Narasimhan2013Structural}
H.~Narasimhan and S.~Agarwal.
\newblock A structural svm based approach for optimizing partial auc.
\newblock In {\em Proc. Int. Conf. Mach. Learn.}, 2013.

\bibitem{Pham2008Detection}
M.-T. Pham, V.-D.~D. Hoang, and T.-J. Cham.
\newblock Detection with multi-exit asymmetric boosting.
\newblock In {\em Proc. IEEE Conf. Comp. Vis. Patt. Recogn.}, 2008.

\bibitem{Qi2006Evaluation}
Y.~Qi, Z.~Bar-Joseph, and J.~Klein-Seetharaman.
\newblock Evaluation of different biological data and computational
  classification methods for use in protein interaction prediction.
\newblock {\em Proteins: Struct., Func., and Bioinfor.}, 63(3):490--500, 2006.

\bibitem{Sermanet2013Pedestrian}
P.~Sermanet, K.~Kavukcuoglu, S.~Chintala, and Y.~LeCun.
\newblock Pedestrian detection with unsupervised multi-stage feature learning.
\newblock In {\em Proc. IEEE Conf. Comp. Vis. Patt. Recogn.}, 2013.

\bibitem{FisherBoost2013IJCV}
C.~Shen, P.~Wang, S.~Paisitkriangkrai, and A.~{van den Hengel}.
\newblock Training effective node classifiers for cascade classification.
\newblock {\em Int. J. Computer Vision}, 103(3):326--347, 2013.

\bibitem{Tuzel2008Pedestrian}
O.~Tuzel, F.~Porikli, and P.~Meer.
\newblock Pedestrian detection via classification on {R}iemannian manifolds.
\newblock {\em {IEEE} Trans. Pattern Anal. Mach. Intell.}, 30(10):1713--1727,
  2008.

\bibitem{Viola2002Fast}
P.~Viola and M.~Jones.
\newblock Fast and robust classification using asymmetric {AdaBoost} and a
  detector cascade.
\newblock In {\em Proc. Adv. Neural Inf. Process. Syst.}, pages 1311--1318. MIT
  Press, 2002.

\bibitem{Viola2004Robust}
P.~Viola and M.~J. Jones.
\newblock Robust real-time face detection.
\newblock {\em Int. J. Comp. Vis.}, 57(2):137--154, 2004.

\bibitem{Walk2010New}
S.~Walk, N.~Majer, K.~Schindler, and B.~Schiele.
\newblock New features and insights for pedestrian detection.
\newblock In {\em Proc. IEEE Conf. Comp. Vis. Patt. Recogn.}, San Francisco,
  US, 2010.

\bibitem{AsymBoost2011Wang}
P.~Wang, C.~Shen, N.~Barnes, and H.~Zheng.
\newblock Fast and robust object detection using asymmetric totally-corrective
  boosting.
\newblock {\em IEEE Trans. Neural Networks and Learning Systems}, 23(1):33--46,
  2012.

\bibitem{Wang2009HOG}
X.~Wang, T.~X. Han, and S.~Yan.
\newblock An {HOG-LBP} human detector with partial occlusion handling.
\newblock In {\em Proc. IEEE Int. Conf. Comp. Vis.}, 2009.

\bibitem{Wu2008Fast}
J.~Wu, S.~C. Brubaker, M.~D. Mullin, and J.~M. Rehg.
\newblock Fast asymmetric learning for cascade face detection.
\newblock {\em {IEEE} Trans. Pattern Anal. Mach. Intell.}, 30(3):369--382,
  2008.

\bibitem{Wu2008Asymmetric}
S.-H. Wu, K.-P. Lin, C.-M. Chen, and M.-S. Chen.
\newblock Asymmetric support vector machines: low false-positive learning under
  the user tolerance.
\newblock In {\em Proc. of Intl. Conf. on Knowledge Discovery and Data Mining},
  2008.

\end{thebibliography}
}

\end{document}